\documentclass{article}

\usepackage[final]{neurips_2024}

\usepackage[utf8]{inputenc} 
\usepackage[T1]{fontenc}    
\usepackage{hyperref}       
\usepackage{url}            
\usepackage{booktabs}       
\usepackage{amsfonts}       
\usepackage{nicefrac}       
\usepackage{microtype}      

\usepackage{graphicx} 
\usepackage{amsmath}
\usepackage{multirow}
\usepackage[table,xcdraw]{xcolor}
\usepackage{bm}
\usepackage[normalem]{ulem}
\usepackage{bbding}
\usepackage{enumitem}

\usepackage{tcolorbox}
\tcbuselibrary{skins}
\tcbuselibrary{breakable}

\usepackage{pifont}

\useunder{\uline}{\ul}{}

\title{Uni-Med: A Unified Medical Generalist Foundation Model For Multi-Task Learning Via Connector-MoE}

\author{%
  Xun Zhu\textsuperscript{1} \quad Ying Hu\textsuperscript{1} \quad Fanbin Mo\textsuperscript{2} \quad Miao Li\textsuperscript{1, \ding{41}} \quad Ji Wu\textsuperscript{1, 3}  \\
  \textsuperscript{1} Department of Electronic Engineering, Tsinghua University \\
  \textsuperscript{2} School of Artificial Intelligence, Beijing University of Posts and Telecommunications \\
  \textsuperscript{3} College of AI, Tsinghua University \\
  \{zhu-x24, yinghu\_yh\}@mails.tsinghua.edu.cn \quad
   mofanbin@bupt.edu.cn \\
  \{miao-li, wuji\_ee\}@tsinghua.edu.cn \quad \textsuperscript{\ding{41}} corresponding author\\
  }

\begin{document}

\maketitle

\begin{abstract}
  Multi-modal large language models (MLLMs) have shown impressive capabilities as a general-purpose interface for various visual and linguistic tasks.
  However, building a unified MLLM for multi-task learning in the medical field remains a thorny challenge.
  To mitigate the tug-of-war problem of multi-modal multi-task optimization in MLLMs, recent advances primarily focus on improving the LLM components, while neglecting the connector that bridges the gap between modalities.
  In this paper, we introduce Uni-Med, a novel medical generalist foundation model which consists of a universal visual feature extraction module, a connector mixture-of-experts (CMoE) module, and an LLM.
  Benefiting from the proposed CMoE that leverages a well-designed router with a mixture of projection experts at the connector, Uni-Med achieves efficient solution to the tug-of-war problem and can perform six different medical tasks including question answering, visual question answering, report generation, referring expression comprehension, referring expression generation and image classification.
  To the best of our knowledge, Uni-Med is the first effort to tackle multi-task interference at the connector in MLLMs.
  Extensive ablation experiments validate the effectiveness of introducing CMoE under any configuration, with up to an average 8\% performance gains.
  We further provide interpretation analysis of the tug-of-war problem from the perspective of gradient optimization and parameter statistics.
  Compared to previous state-of-the-art medical MLLMs, Uni-Med achieves competitive or superior evaluation metrics on diverse tasks.
  Code and resources are available at \url{https://github.com/tsinghua-msiip/Uni-Med}.
\end{abstract}

\section{Introduction}
Driven by the growth of datasets, the increase in model size, and advances in generative language foundation models \citep{achiam2023gpt,touvron2023llama}, multi-modal large language models (MLLMs) now offer unprecedented abilities as general-purpose interfaces.
These advancements are spurring innovation across various visual and linguistic tasks \citep{chen2023x,lyu2023macaw,su2023pandagpt}.
While significant strides have been made in building a unified foundation model for natural scenery \citep{chen2022unified,lu2022unified,lu2023unified}, the development of generalist medical artificial intelligence is still in its early stages \citep{moor2023foundation}.

The goal of a unified and generalist medical foundation model is to enable joint training on massive medical datasets. 
This model aims to handle multiple tasks and modalities within a single architecture with shared parameters \citep{zhang2023biomedgpt,li2024llava}.
It seeks to eliminate the need for task-specific modules and further fine-tuning, thereby revolutionizing the traditional task-specific approach to model development \citep{wu2023towards,tu2024towards}.
However, existing open-source efforts have not yet fully achieved these ambitious goals.

A key challenge in creating a unified medical foundation model is the complexity of multi-modal, multi-task learning, often exacerbated by the tug-of-war problem \citep{hadsell2020embracing}.
Inherent task conflicts and data imbalances can cause interference during the simultaneous learning of different tasks.
This problem is particularly acute in the medical field, where tasks and modalities are highly specialized and diverse.
As a result, the performance of each task may degrade compared to task-specialized models \citep{yu2020gradient,zhu2022uni}.

\begin{figure}
\centerline{\includegraphics[scale=0.32]{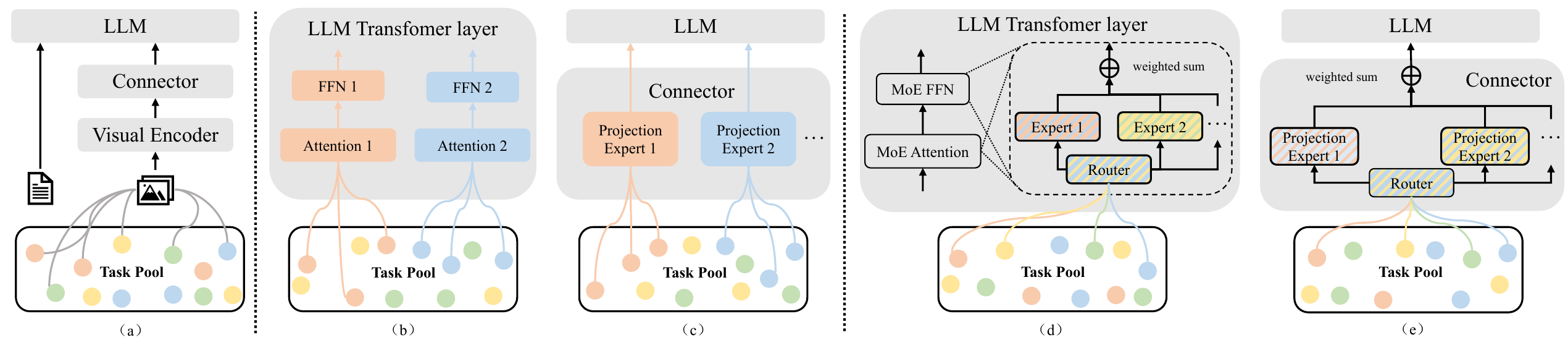}}
\caption{
Three hypotheses and corresponding architectural implementations for multi-task learning in MLLMs.
(a) Synergy hypothesis. (b)-(c) Conflict hypothesis in LLM and connector, respectively. (d)-(e) Conflict-synergy coexist hypothesis in LLM and connector, respectively.
}
\label{F1}
\end{figure}

To mitigate the tug-of-war problem in multi-task learning, recent advances introduce the well-known Mixture-of-Experts (MoE) \citep{jacobs1991adaptive} into MLLMs.
Figure~\ref{F1} illustrates three distinct hypotheses and their corresponding architectural implementations for multi-task learning in MLLMs.
The first "synergy hypothesis" suggests that all tasks benefit from a fully shared backbone comprising a visual encoder, connector, and language model, which is the standard architecture for MLLMs.
The second "conflict hypothesis", proposes that each task requires its own specific adaptations, thereby preventing knowledge sharing among tasks.
The third "conflict-synergy coexistence hypothesis", posits that all tasks share multi-task adaptations, which reduces interference and promotes more efficient knowledge sharing. 
However, current research \citep{zadouri2023pushing,gou2023mixture,liu2023moelora,lin2024moe} mainly tailors the MoE approach to the language model components, overlooking the potential benefits of exploring and enhancing the connector in MLLMs.
Furthermore, the optimization of the tug-of-war problem lacks a detailed, explainable analysis.

In this study, we first identify a tug-of-war problem in multi-task learning at the connector level within standard MLLM architectures.
This issue indicates that different tasks may emphasize different types of features in multi-modal, multi-task scenarios. Consequently, a fully shared connector may fall short as it cannot accommodate the diverse modal features required by each task.
Drawing inspiration from the successful application of MoE in LLMs, we introduce Connector-MoE (CMoE), a novel approach that employs a mixture of projection experts to align visual and language embedding spaces effectively, thus mitigating the tug-of-war problem.
As a pioneering effort in constructing a unified generalist foundation model in the medical field, we present Uni-Med.
This model comprises a universal visual feature extraction module, a CMoE module, and an LMM.
Uni-Med demonstrates impressive performance across six distinct medical tasks, with minimal training computational overhead.
It achieves joint training on 12 datasets on a single A800 in under 10 hours.
The effectiveness and generalization of CMoE are underscored through ablation experiments. 
Additionally, an interpretable analysis reveals that Uni-Med provides a superior solution to the tug-of-war problem at the connector level.
Overall, Uni-Med delivers competitive or even superior performance compared to open-source, state-of-the-art medical MLLMs on all test sets.
Our contributions can be summarized as:
\begin{itemize}
\item We present Uni-Med, an open-source medical generalist foundation model with a unified interface and shared parameters, which can perform six different medical tasks including question answering, visual question answering, report generation, referring expression comprehension, referring expression generation and image classification.
\item We propose CMoE, a well-designed connector component for MLLMs, which significantly outperforms baselines under any configuration, with up to an average 8\% performance gains. To our knowledge, Uni-Med is the first attempt to focus on the connector in MLLMs to mitigate the tug-of-war problem, which is critical but has always been overlooked.
\item Focusing on the question of how the tug-of-war problem is optimized, which has never been quantitatively discussed, we provide detailed interpretability analysis and instructive findings from the perspective of gradient optimization and parameter statistics.
\item Uni-Med achieves competitive or superior performance compared to the open-source, state-of-the-art medical MLLMs on test set of diverse tasks and datasets, which demonstrates the huge potential of medical generalist foundation models.
\end{itemize}

\section{Related work}
\paragraph{Medical foundation models}
The increasing availability of medical data, as well as advances in multi-modal LLM technologies, have paved the way for the emergence of medical foundational models.
Med-Flamingo \citep{moor2023med} continues pre-training on paired and interleaved medical image-text data based on OpenFlamingo \citep{awadalla2023openflamingo}. 
LLaVA-Med \citep{li2024llava} curates a medical multi-modal instruction following dataset and fine-tunes LLaVA \citep{liu2024visual} with it. 
XrayGPT \citep{thawkar2023xraygpt} can analyze and answer open-ended questions about chest X-rays. 
BiomedGPT \citep{zhang2023biomedgpt} is a multi-task foundation model pretrained on a diverse source of medical images, literature, and clinical notes.
However, most of these efforts require further fine-tuning on task-specific data to support downstream applications.
One step further, the generalist foundation model uses the same weight to excel at various tasks without fine-tuning.
RadFM \citep{wu2023towards} is dedicated to build a generalist foundation model for radiology.
Med-PaLM M \citep{tu2024towards} is directly trained in a unified framework to jointly handle many tasks, which is perhaps most similar to our effort, but it does not provide access for usage.
In addition, recent studies \citep{wu2023can, yan2024worse, xia2024cares} have suggested the the necessity for a more comprehensive and detailed evaluation of the capabilities of medical MLLMs.

\paragraph{MoE in multi-task learning}
MoE is originally considered to increase the model capacity \citep{riquelme2021scaling,fedus2022switch} and gains popularity in mitigating multi-task interference \citep{chen2023mod,chen2024llava}. It achieves this by utilizing a router to determine the token set handled by each expert, thus reducing interference between different types of samples.
Recent studies have focused on combining MoE with LLM, such as MoE-LLaVA \citep{lin2024moe} and Mixtral 8x7B \citep{jiang2024mixtral}, or combining MoE with one of the representative parameter-efficient tuning techniques, i.e., LoRA \citep{hu2021lora}, such as Octavius \citep{chen2023octavius}, MoCLE \citep{gou2023mixture}, MTLoRA \citep{agiza2024mtlora} and MOELoRA\citep{liu2024moe}.
However, neither of them introduces MoE into the connector component for MLLMs.
Furthermore, there is a lack of clear and explicit interpretable analysis on how  the multi-task interference is mitigated through the use of MoE.

\paragraph{Cross-modality connector in MLLM}
The connector between the multi-modal encoder and the LLM is critical in aligning multi-modal features \citep{song2023bridge}.
One of the most popular paradigms is to map multi-modal features into a feature space that aligns with language, such as linear projection \citep{liu2024visual} and MLP projection \citep{liu2023improved,chen2023sharegpt4v}.
Another paradigm is to transform multi-modal features into multi-modal tokens that are consistent with the embedded representation space of LLM, such as cross-attention \citep{li2022blip,ye2023mplug,ye2024mplug}, perceiver resampler \citep{alayrac2022flamingo,peng2023kosmos} and
Q-Former \citep{li2023blip,zhu2023minigpt}.
However, existing paradigms use the same connector when processing the same modal data for different tasks, ignoring the imperative to acquire distinct alignment patterns tailored to the demands of each task.

\section{Methodology}
\subsection{Preliminaries}
\subsubsection{Multi-task interference}
\label{3.1.1}
To quantify the intricate tug-of-war problem in a unified foundation model, we provide interpretability from the perspective of gradient optimization and parameter statistics.
\paragraph{Perspective of gradient optimization}
When optimizing the shared parameters $\theta$ according to task $j$,
the change in the update direction of loss $L_i$ for task $i$ can be defined as
\citep{zhu2022uni}:
\vspace{-1pt}
\begin{small}
\begin{equation}
\Delta_j L_i\left(x_i\right) \doteq \mathbb{E}_{x_j}\left(L_i\left(x_i ; \theta\right)-L_i\left(x_i ; \theta-\lambda \frac{\nabla_\theta L_j\left(x_j\right)}{\left\|\nabla_\theta L_j\left(x_j\right)\right\|_2}\right)\right) \approx \lambda \mathbb{E}_{x_j}\left(\frac{\nabla_\theta L_j\left(x_j\right)^T}{\left\|\nabla_\theta L_j\left(x_j\right)\right\|_2} \nabla_\theta L_i\left(x_i\right)\right)
\end{equation}
\end{small}

where $x_i$ and $x_j$ are the sampled training batches of task $i$ and $j$, respectively.
The interference of task $j$ on task $i$ in the update direction can be quantified as:
\begin{small}
\begin{equation}
\mathcal{GD}_{i, j}=\mathbb{E}_{x_i}\left(\frac{\Delta_j L_i\left(x_i\right)}{\Delta_i L_i\left(x_i\right)}\right)
\end{equation}
\end{small}

The gradient magnitude similarity between task $i$ and task $j$ can be defined as:
\begin{small}
\begin{equation}
\mathcal{GM}_{i, j}=\mathcal{GM}_{j, i}=\frac{2\mathbb{E}_{x_i}\left(\left\|\nabla_\theta L_i\left(x_i\right)\right\|_2\right)\mathbb{E}_{x_j}\left(\left\|\nabla_\theta L_j\left(x_j\right)\right\|_2\right)}{\left(\mathbb{E}_{x_i}\left(\left\|\nabla_\theta L_i\left(x_i\right)\right\|_2\right)\right)^2
+\left(\mathbb{E}_{x_j}\left(\left\|\nabla_\theta L_j\left(x_j\right)\right\|_2\right)\right)^2}
\end{equation}
\end{small}

$\mathcal{GM}_{i, j}$ goes to zero when the difference in gradient magnitudes is large, indicating that some task is dominant \citep{yu2020gradient}.
For all $T$ tasks, we can get $\bm{\mathcal{GD}},\bm{\mathcal{GM}} \in \mathbf{R}^{{T}\times T}$.
Then, we define the tug-of-war indexes for each task in multi-task learning through the function $G$ as follows:
\begin{small}
\begin{equation}
\textit{tug-of-war indexes} = G(\bm{\mathcal{GD}},\bm{\mathcal{GM}})=\left[\sum\nolimits_{j=1}^T \mathcal{GD}_{i, j} \cdot \mathcal{GM}_{i, j}\right]_{i=1}^T
\end{equation}
\end{small}

\paragraph{Perspective of parameter statistics}
Inspired by the Gradient Positive Sign Purity proposed by \citet{chen2020just}, we define the statistics score of a single parameter in multi-task learning:
\begin{small}
\begin{equation}
\textit{statistics score}=\left|\frac{\sum_i^T \nabla_\theta L_i}{\sum_i^T\left|\nabla_\theta L_i\right|}\right|
\end{equation}
\end{small}

where $\nabla_\theta L_i$ is the gradient for task $i$. 
The range of the statistics score is [0, 1], and a value close to 1 indicates that this parameter suffers less gradient conflict during multi-task training.
Upon collecting the statistics scores of all parameters, we can intuitively demonstrate and analyze the phenomenon of multi-task interference.

\begin{figure}
  \centerline{\includegraphics[scale=0.51]{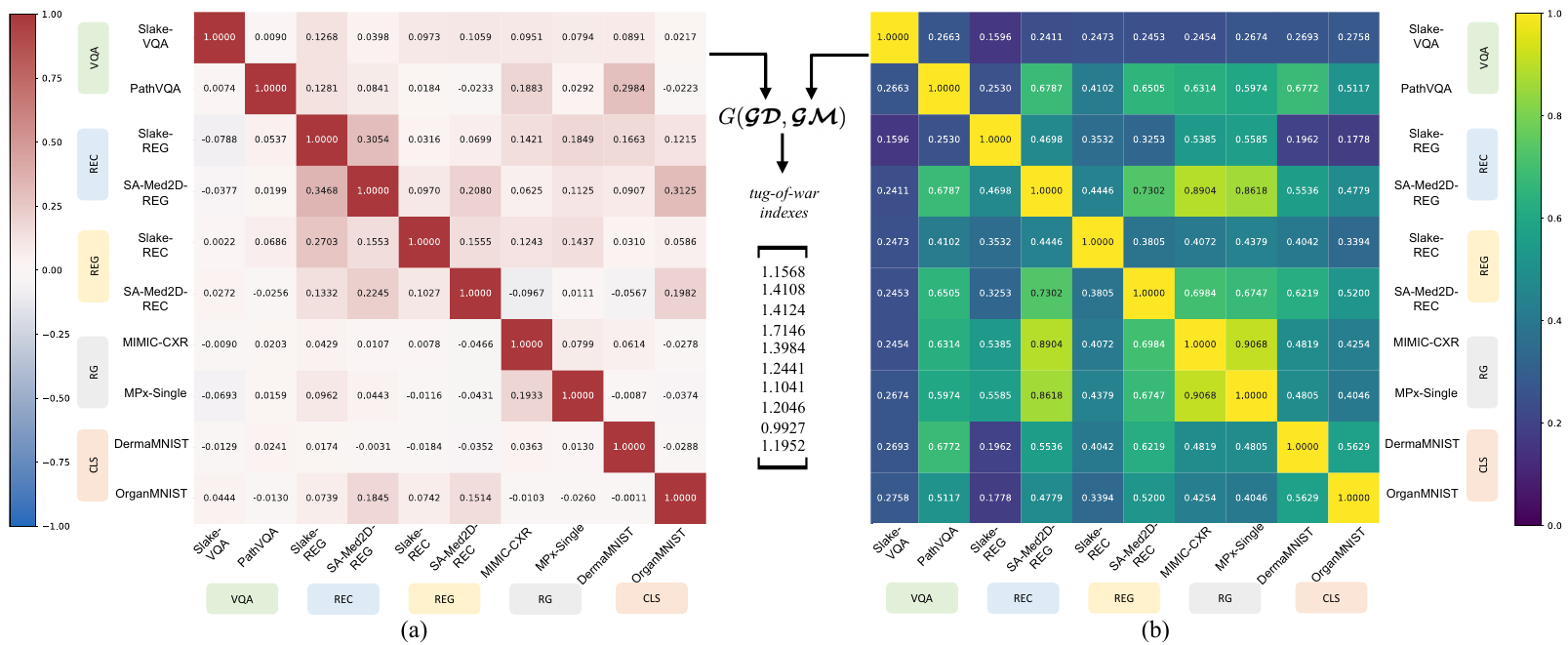}}
  \caption{Dataset-level multi-task interference of the synergy hypothesis model at the connector in MLLMs. (a) Perspective of gradient direction $\bm{\mathcal{GD}}$. (b) Perspective of gradient magnitude $\bm{\mathcal{GM}}$.}
\label{F2}
\end{figure}

To be specific, we sample 100 batches for each datasets and record the gradients to calculate all of the above metrics.
Figure~\ref{F2} shows the dataset-level (more granular than task-level) multi-task interference of the synergy hypothesis model at the connector in the standard MLLM architecture. 

\subsubsection{Mixture-of-Experts}
A Mixture-of-Experts (MoE) contains a set of expert networks $E_1, E_2, ..., E_N$ along with a routing network $R$. 
For each token $x_i$ in the input sequence $\bm{X}=\{x_i\}_{i=1}^L$, the output of MoE is the weighted sum of outputs from each expert, where the weight is calculated by the router:
\begin{small}
\begin{equation}
y_i=\sum_{k=1}^N R(x_i)_k \cdot E_k(x_i)
\end{equation}
\end{small}

The types of $R$ can mainly be divided into: 
1) Constant router, which assigns equal weight to each expert.
2) Hard router, which enforces one-to-one mapping between tasks and experts.
3) Sparse router, which selects Top-K experts with the maximum routing weight.
4) Soft router, which calculates the routing weights for each expert.
For more details on the routing networks, see Appendix~\ref{A.1}.
\subsection{Model Architecture}
\begin{figure}
  \centerline{\includegraphics[scale=0.36]{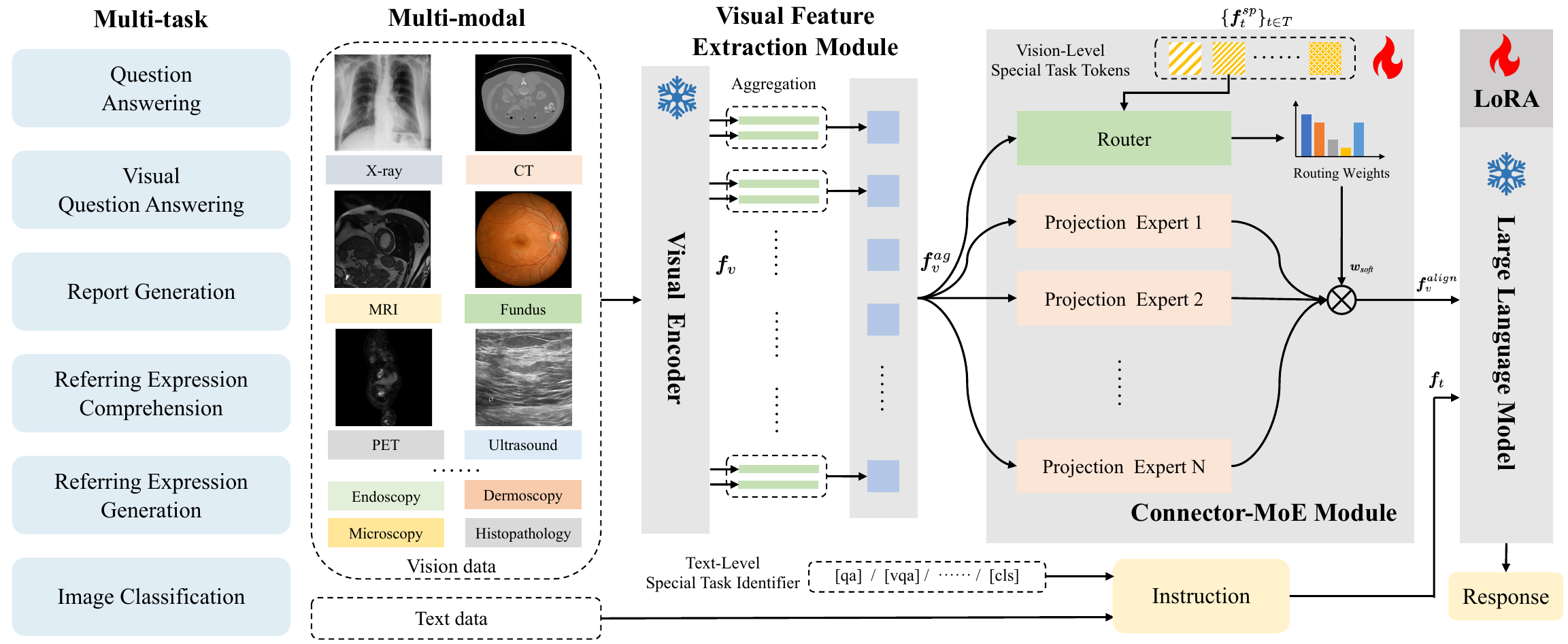}}
  \caption{Overall architecture of Uni-Med, which consists of a universal vision feature extraction module, a connector-MoE module and an LLM. Uni-Med can perform six different medical tasks including question answering, visual question answering, report generation, referring expression comprehension, referring expression generation and image classification.}
\label{F3}
\end{figure}
With the primary goal of achieving a unified medical generalist foundation
model and mitigating the tug-of-war problem of multi-task learning in mind,
we design the overall architecture of Uni-Med as illustrated in Figure~\ref{F3}, which contains three components:
a universal vision feature extraction module, a connector-MoE module and an LLM.
Detailed descriptions are presented in the following sections.

\subsubsection{Visual feature extraction module}
Taking one of the multi-modal medical images $\bm{I} \in \mathbf{R}^{{H}\times W\times C}$ as input, the visual encoder $\bm{V}_{en}$ extracts the image tokens $\bm{f}_v \in \mathbf{R}^{{N_v}\times {D_v}}$ for image perception, where ${N_v} = HW /{P^{2}}$ is the number of image patches and ${D_v}$ is the hidden size of visual embeddings.

To alleviate the efficiency issues caused by prolonged visual input tokens during the training and inference, we scheme a resampler with a compression rate $\alpha$ for visual feature aggregation.
Concretely, $\alpha$ adjacent visual tokens are concatenated and projected into one single embedding.
Thus we obtain aggregated image tokens $\bm{f}_{v}^{ag} \in \mathbf{R}^{{N_v / \alpha}\times {D_v \alpha}}$ as follows:
\begin{small}
\begin{equation}
\bm{f}_{v}^{ag}=\textit{resampler}\left(\bm{V}_{en}\left( \bm{I} \right), \alpha\right)
\end{equation}
\end{small}

\subsubsection{Connector-MoE module}
Aligning the visual space with the language embedding space of the large language model is a critical process, especially in the complex and diverse input of multi-task multi-modal medical image text pairs.
Based on the conflict-synergy coexist hypothesis, we propose the Connector-MoE (CMoE) module, which aims to adaptively minimize task conflict and maximize task synergy at the connector.
CMoE module has $N$ projection experts $E_1, E_2, ..., E_N$, where each expert is a two-layer MLP, and a soft router $R_{\textit{soft}}$ to control the contribution of each expert.

According to Figure~\ref{F2},
we find that: 
(1) Gradient optimization conflict is common and consistent at the task level.
(2) Even for the same task, there are significant differences in conflict and synergy at dataset-level.
To alleviate the above problems, we randomly initialize vision-level special task tokens $\{\bm{f}_{t}^{sp} \}_{t \in T}$, where $\bm{f}_{t}^{sp} \in \mathbf{R}^{{D_v \alpha}}$ and $T$ is the set of tasks.
$R_{\textit{soft}}$ is a lightweight MLP designed to receive the concatenated inputs of $\bm{f}_{v}^{ag}$ (token level) and $\bm{f}_{t}^{sp}$ (task level), and calculate the routing weights $\bm{w}_{\textit{soft}}\in \mathbf{R}^{{D_v / \alpha} \times {N}}$ of each expert for each image token, which can be formulated as:
\vspace{-1pt}
\begin{small}
\begin{equation}
\bm{w}_{\textit{soft}}\left(\bm{f}_{v}^{ag}\right)=
\sigma \cdot
R_{\textit{soft}}\left(
\left[\bm{f}_{v}^{ag}, Repeat \left( \bm{f}_{t}^{sp}\right)\right]\right)
\end{equation}
\end{small}

where $\left[,\right]$ denotes concatenation operation, $\sigma$ is \textit{SoftMax} function.
Then we can obtain aligned visual tokens $\bm{f}_{v}^{align} \in \mathbf{R}^{{N_v / \alpha}\times {D_t}}$ through a weighted sum of all experts' output as follows:
\begin{small}
\begin{equation}
\setlength{\abovedisplayskip}{3pt}
\setlength{\belowdisplayskip}{3pt}
\bm{f}_{v}^{align}=\sum_{k=1}^N \bm{w}_{\textit{soft},k} \cdot E_k(\bm{f}_{v}^{ag})
\end{equation}
\end{small}

where ${D_t}$ is the hidden size of the language embedding space of the large language model and $\bm{w}_{\textit{soft},k}$ denotes the routing weight of the $k$-th projection expert.
We discuss and analyze the effects of router type, router strategy, and number of experts in Section~\ref{4.2.1}.

\begin{table}
\scriptsize
\renewcommand\arraystretch{0.9}
\tabcolsep= 0.3cm
  \caption{Text-level special task identifiers for different tasks.}
  \label{T1}
  \centering
\begin{tabular}{c c c c c c c}
\toprule
\multirow{3}{*}{\textbf{Task}} & \multirow{3}{*}{\begin{tabular}[c]{@{}c@{}}Question\\ Answering\end{tabular}} & \multirow{3}{*}{\begin{tabular}[c]{@{}c@{}}Visual\\ Question\\ Answering\end{tabular}} & \multirow{3}{*}{\begin{tabular}[c]{@{}c@{}}Report\\ Generation\end{tabular}} & \multirow{3}{*}{\begin{tabular}[c]{@{}c@{}}Referring\\ Expression \\ Comprehension\end{tabular}} & \multirow{3}{*}{\begin{tabular}[c]{@{}c@{}}Referring\\ Expression\\ Generation\end{tabular}} & \multirow{3}{*}{\begin{tabular}[c]{@{}c@{}}Image\\ Classification\end{tabular}} \\
 &  &  &  &  &  &  \\
 &  &  &  &  &  &  \\ \midrule
\textbf{Identifier} & {[}qa{]} & {[}vqa{]} & {[}caption{]} & {[}refer{]} & {[}identify{]} & {[}cls{]} \\
\bottomrule
\end{tabular}
\end{table}

\subsubsection{Large language model}
Similar to the vision-level special task tokens, we assign the text-level special task identifiers for question answering (QA), visual question answering (VQA), report generation (RG), referring expression comprehension (REC), referring expression generation (REG) and image classification (CLS) as shown in Table~\ref{T1}, which can help reduce multi-task ambiguity \citep{chen2023minigpt}.
The text prompt is designed as
"<Img> < ImageFeature> </Img> [Task Identifier] Instruction",
which merges the converted image features with the textual instructions.
See details about our multi-task instruction template in Appendix~\ref{C}.

After word embedding, we can obtain textual tokens $\bm{f}_t \in \mathbf{R}^{{N_t}\times {D_t}}$, where $N_t$ denotes the number of textual tokens.
LLM generates the response $\bm{O} = \{O_i\}_{i=1}^{L}$ conditioned on the aligned visual tokens $\bm{f}_{v}^{align}$ and textual tokens $\bm{f}_t$ inputs in an autoregressive manner, which can be formulated as:
\begin{small}
\begin{equation}
\setlength{\abovedisplayskip}{3pt}
\setlength{\belowdisplayskip}{3pt}
p\left(\mathbf{O}_t \mid \bm{f}_{v}^{align}, \bm{f}_t\right)=\prod_{i=1}^L p\left(O_i \mid \bm{f}_{v}^{align}, \bm{f}_t, O_{<i}\right)
\end{equation}
\end{small}

where $L$ is the length of output tokens.
We use low-rank adaption (LoRA) \citep{hu2021lora} for efficient LLM fine-tuning, which is applied to all the linear layers.

\section{Experiments}
\subsection{Experiment settings}
\paragraph{Tasks and datasets}
Text-only data is collected from MedQA \citep{jin2021disease} and PubMedQA \citep{jin2019pubmedqa} for the task of QA.
Image-text pairs are collected from Path-VQA \citep{he2020pathvqa} and Slake-VQA \citep{liu2021slake} for the task of VQA, MIMIC-CXR \citep{johnson2019mimic} and MPx-Single \citep{wu2023towards} for the task of RG, MedMNIST v2 \citep{yang2023medmnist} for the task of CLS.
For tasks such as REG and REC that require representation of spatial locations, we use the bounding boxes of the format "<$X_{min}$><$Y_{min}$><$X_{max}$><$Y_{max}$>", which denotes the coordinates of objects.
Then, we respectively process datasets Slake-VQA \citep{liu2021slake} and SA-Med2D-20M \citep{ye2023sa} to get datasets Slake-REC, Slake-REG, SA-Med2D-REC, and SA-Med2D-REG.
For a detailed description, processing and splitting of all datasets, see Appendix~\ref{B}.

\paragraph{Implementation details}
We adapt the open-sourced ViT-G/14 from EVA-CLIP \citep{fang2023eva} and LLaMA2-Chat (7B) \citep{touvron2023llama} as our visual backbone and LLM, respectively. 
During the training process, each task is assigned a sample rate that is calculated in proportion to the respective task's data volume.
The visual backbone remains frozen with an input image resolution of 224*224 and the LLM is fine-tuned through LoRA \citep{hu2021lora} with the rank of 8.
The compression rate $\alpha$=4 and the number of projection experts $N$=5.
Uni-Med only requires one-stage training on a NVIDIA A800-SXM4-80GB GPU, with the first 10k iterations to warm-up and a total of 100k iterations with a batch size of 4, which lasts roughly 10 hours.
The peak learning rate is set to 1e-6 and it decays to 1e-7 following the cosine strategy.
We use AdamW \citep{loshchilov2017decoupled} optimizer with $\beta_{1}$=0.9, $\beta_{2}$=0.95 and weight decay of 0.05.

\begin{table}[!b]
\tiny
\tabcolsep= 0.056cm
\caption{Experiments of ablation study.
Metrics are reported on "Slake-VQA/Path-VQA", "Slake-REC/SA-Med2D-REC", "Slake-REG/SA-Med2D-REG", "MIMIC-CXR/MPx-Single", "DermaMNIST/OrganSMNIST" for the task of VQA, REC, REG, RG, and CLS, respectively.}
\label{T2}
\begin{tabular}{cccccccccccccc}
\toprule
& \multicolumn{2}{c}{\textbf{Router}} & \textbf{VQA} &  & \textbf{REC} &  & \textbf{REG}&  & \textbf{RG} &  & \textbf{CLS} &  &  \\
\multirow{-2}{*}{\textbf{Connector}} & \textbf{Type} & \textbf{Strategy} & BLEU-1 & \multirow{-2}{*}{$\Delta \left(\uparrow\right)$} & IoU & \multirow{-2}{*}{$\Delta \left(\uparrow\right)$} & BLEU-1 & \multirow{-2}{*}{$\Delta \left(\uparrow\right)$} & BLEU-1 & \multirow{-2}{*}{$\Delta \left(\uparrow\right)$} & Accuracy & \multirow{-2}{*}{$\Delta \left(\uparrow\right)$} & \multirow{-2}{*}{\begin{tabular}[c]{@{}c@{}}\textbf{Total}\\ \textbf{$\Delta \left(\uparrow\right)$}\end{tabular}} \\

\midrule
\multicolumn{14}{l}{\textbf{(a) Connector design}} \\
\midrule

Linear & - & - & 77.90 / 56.27 & -1.4\% & 28.44 / 11.59 & -23.9\% & 74.98 / 55.61 & -2.1\% & 13.80 / 15.85 & -11.6\% & 72.47 / 69.39 & -5.4\% & -8.9\% \\
MLP & - & - & \cellcolor[HTML]{F2F2F2}79.81 / 56.48 & \cellcolor[HTML]{F2F2F2} & \cellcolor[HTML]{F2F2F2}35.18 / 16.26 & \cellcolor[HTML]{F2F2F2} & \cellcolor[HTML]{F2F2F2}74.54 / 58.42 & \cellcolor[HTML]{F2F2F2} & \cellcolor[HTML]{F2F2F2}18.55 / 15.50 & \cellcolor[HTML]{F2F2F2} & \cellcolor[HTML]{F2F2F2}76.26 / 73.64 & \cellcolor[HTML]{F2F2F2} & \cellcolor[HTML]{F2F2F2} \\
 & Constant & - & 82.74 / 57.38 & 2.6\% & 33.94 / 15.49 & -4.1\% & 73.58 / 58.51 & -0.6\% & 23.16 / 15.88 & 13.7\% & 75.91 / 76.50 & 1.7\% & 2.7\% \\
 & Hard & - & 81.85 / \textbf{59.09} & \textbf{3.6\%} & 30.01 / 11.59 & -21.7\% & 70.91 / 58.04 & -2.8\% & 22.76 / 15.79 & 12.3\% & \textbf{81.55} / \textbf{81.18} & \textbf{8.6\%} & 0.0\% \\
 & Sparse & Token & 80.68 / 57.02 & 1.0\% & 37.07 / 18.41 & 9.3\% & 76.86 / 60.08 & 3.0\% & 24.02 / 15.73 & 15.5\% & 73.47 / 74.93 & -1.0\% & 5.6\% \\
 &  & Token & 81.79 / 57.69 & 2.3\% & 35.51 / 17.79 & 5.2\% & 74.43 / \textbf{61.34} & 2.4\% & \textbf{26.27} / 15.61 & \textbf{21.2\%} & 76.56 / 77.21 & 2.6\% & 6.7\% \\
 &  & Task & \textbf{82.51} / 57.43 & 2.5\% & \textbf{38.33} / 19.68 & 15.0\% & \textbf{78.18} / 60.67 & \textbf{4.4\%} & 23.34 / \textbf{15.89} & 14.2\% & 77.56 / 76.55 & 2.8\% & 7.8\% \\
\multirow{-6}{*}{CMoE} & \multirow{-3}{*}{Soft} & Token\&Task & \cellcolor[HTML]{D9D9D9}81.52 / 57.75 & \cellcolor[HTML]{D9D9D9}2.2\% & \cellcolor[HTML]{D9D9D9}37.54 / \textbf{20.30} & \cellcolor[HTML]{D9D9D9}\textbf{15.8\%} & \cellcolor[HTML]{D9D9D9}77.45 / 60.42 & \cellcolor[HTML]{D9D9D9}3.7\% & \cellcolor[HTML]{D9D9D9}24.70 / 15.55 & \cellcolor[HTML]{D9D9D9}16.7\% & \cellcolor[HTML]{D9D9D9}75.61 / 76.92 & \cellcolor[HTML]{D9D9D9}1.8\% & \cellcolor[HTML]{D9D9D9}\textbf{8.0\%} \\

\midrule
\multicolumn{14}{l}{\textbf{(b) Resampler design}} \\
\midrule

\multicolumn{3}{l}{Compression Rate = 1} & 79.63 / \textbf{58.34} & 1.5\% & 30.20 / 14.48 & -12.6\% & 70.81 / 60.12 & -1.0\% & 23.92 / 15.54 & 14.6\% & \textbf{78.20} / 76.16 & \textbf{3.0\%} & 1.1\% \\
\multicolumn{3}{l}{Compression Rate = 2, Projection} & \textbf{83.74} / 57.70 & \textbf{3.5\%} & 37.02 / 18.57 & 9.7\% & 71.89 / 60.32 & -0.2\% & \textbf{25.83} / 15.77 & \textbf{20.5\%} & 74.56 / 76.70 & 1.0\% & 6.9\% \\
\multicolumn{3}{l}{Compression Rate = 4, Max Pooling} & 80.36 / 57.44 & 1.2\% & 27.16 / 14.37 & -17.2\% & 68.30 / 57.56 & -4.9\% & 18.85 / 15.60 & 1.1\% & 75.71 / 73.08 & -0.7\% & -4.1\% \\
\multicolumn{3}{l}{Compression Rate = 4, Avg Pooling} & 81.96 / 57.93 & 2.6\% & 34.21 / 14.76 & -6.0\% & 73.39 / 59.59 & 0.2\% & 22.18 / \textbf{15.88} & 11.0\% & 72.42 / 74.54 & -1.9\% & 1.2\% \\
\multicolumn{3}{l}{Compression Rate = 4, Projection} & \cellcolor[HTML]{D9D9D9}81.52 / 57.75 & \cellcolor[HTML]{D9D9D9}2.2\% & \cellcolor[HTML]{D9D9D9}\textbf{37.54} / \textbf{20.30} & \cellcolor[HTML]{D9D9D9}\textbf{15.8\%} & \cellcolor[HTML]{D9D9D9}\textbf{77.45} / \textbf{60.42} & \cellcolor[HTML]{D9D9D9}\textbf{3.7\%} & \cellcolor[HTML]{D9D9D9}24.70 / 15.55 & \cellcolor[HTML]{D9D9D9}16.7\% & \cellcolor[HTML]{D9D9D9}75.61 / \textbf{76.92} & \cellcolor[HTML]{D9D9D9}1.8\% & \cellcolor[HTML]{D9D9D9}\textbf{8.0\%} \\

\midrule
\multicolumn{14}{l}{\textbf{(c) Number of projection experts}} \\
\midrule

\multicolumn{3}{c}{3} & 80.45 / 56.88 & 0.8\% & 35.98 / 17.36 & 4.5\% & 66.64 / 58.10 & -5.6\% & 24.00 / 16.00 & 16.3\% & 74.86 / 74.59 & -0.3\% & 3.1\% \\
\multicolumn{3}{c}{5} & \cellcolor[HTML]{D9D9D9}81.52 / 57.75 & \cellcolor[HTML]{D9D9D9}2.2\% & \cellcolor[HTML]{D9D9D9}37.54 / \textbf{20.30} & \cellcolor[HTML]{D9D9D9}\textbf{15.8\%} & \cellcolor[HTML]{D9D9D9}\textbf{77.45} / 60.42 & \cellcolor[HTML]{D9D9D9}3.7\% & \cellcolor[HTML]{D9D9D9}24.70 / 15.55 & \cellcolor[HTML]{D9D9D9}16.7\% & \cellcolor[HTML]{D9D9D9}75.61 / 76.92 & \cellcolor[HTML]{D9D9D9}1.8\% & \cellcolor[HTML]{D9D9D9}\textbf{8.0\%} \\
\multicolumn{3}{c}{8} & 82.71 / 57.86 & 3.0\% & 36.66 / 18.34 & 8.5\% & 71.15 / 58.40 & -2.3\% & 24.47 / \textbf{15.74} & 16.7\% & \textbf{77.91} / 76.53 & 3.0\% & 5.8\% \\
\multicolumn{3}{c}{10} & \textbf{83.21} / 57.85 & 3.3\% & \textbf{38.70} / 19.01 & 13.5\% & 75.06 / 61.43 & 2.9\% & 25.02 / 15.05 & 16.0\% & 77.66 / \textbf{77.99} & \textbf{3.9\%} & 7.9\% \\
\multicolumn{3}{c}{16} & 82.92 / \textbf{58.70} & \textbf{3.9\%} & 35.74 / 17.66 & 5.1\% & 76.74 / \textbf{61.45} & \textbf{4.1\%} & \textbf{27.18} / 15.48 & \textbf{23.2\%} & 75.86 / 77.36 & 2.3\% & 7.7\% \\

\midrule
\multicolumn{14}{l}{\textbf{(d) Module generalization under LoRA rank setting}} \\
\midrule

rank & \multicolumn{2}{c}{Connector\&Router} &  &  &  &  &  &  &  &  &  & &  \\
 & \multicolumn{2}{c}{MLP} & \cellcolor[HTML]{F2F2F2}80.70 / 56.42 & \cellcolor[HTML]{F2F2F2} & \cellcolor[HTML]{F2F2F2}36.35 / 16.32 & \cellcolor[HTML]{F2F2F2} & \cellcolor[HTML]{F2F2F2}64.34 / 57.02 & \cellcolor[HTML]{F2F2F2} & \cellcolor[HTML]{F2F2F2}22.70 / 15.54 & \cellcolor[HTML]{F2F2F2} & \cellcolor[HTML]{F2F2F2}71.62 / 74.06 & \cellcolor[HTML]{F2F2F2} & \cellcolor[HTML]{F2F2F2} \\
 & \multicolumn{2}{c}{CMoE,Hard} & 81.94 / 57.77 & 2.0\% & 29.30 / 10.98 & -26.1\% & 70.14 / 51.45 & -0.4\% & 22.74 / 15.76 & 0.8\% & 81.95 / 80.46 & 11.5\% & -2.4\% \\
\multirow{-3}{*}{4} & \multicolumn{2}{c}{CMoE,Soft} & \cellcolor[HTML]{D9D9D9}\textbf{82.63} / 57.66 & \cellcolor[HTML]{D9D9D9}2.3\% & \cellcolor[HTML]{D9D9D9}32.80 / 15.31 & \cellcolor[HTML]{D9D9D9}-8.0\% & \cellcolor[HTML]{D9D9D9}68.12 / \textbf{60.84} & \cellcolor[HTML]{D9D9D9}\textbf{6.3\%} & \cellcolor[HTML]{D9D9D9}24.46 / 15.61 & \cellcolor[HTML]{D9D9D9}4.1\% & \cellcolor[HTML]{D9D9D9}76.11 / 73.84 & \cellcolor[HTML]{D9D9D9}3.0\% & \cellcolor[HTML]{D9D9D9}1.5\% \\
 & \multicolumn{2}{c}{MLP} & \cellcolor[HTML]{F2F2F2}79.81 / 56.48 & \cellcolor[HTML]{F2F2F2} & \cellcolor[HTML]{F2F2F2}35.18 / 16.26 & \cellcolor[HTML]{F2F2F2} & \cellcolor[HTML]{F2F2F2}74.54 / 58.42 & \cellcolor[HTML]{F2F2F2} & \cellcolor[HTML]{F2F2F2}18.55 / 15.50 & \cellcolor[HTML]{F2F2F2} & \cellcolor[HTML]{F2F2F2}76.26 / 73.64 & \cellcolor[HTML]{F2F2F2} & \cellcolor[HTML]{F2F2F2} \\
 & \multicolumn{2}{c}{CMoE,Hard} & 81.85 / \textbf{59.09} & 3.6\% & 30.01 / 11.59 & -21.7\% & 70.91 / 58.04 & -2.8\% & 22.76 / 15.79 & 12.3\% & 81.55 / \textbf{81.18} & 8.6\% & 0.0\% \\
\multirow{-3}{*}{8} & \multicolumn{2}{c}{CMoE,Soft} & \cellcolor[HTML]{D9D9D9}81.52 / 57.75 & \cellcolor[HTML]{D9D9D9}2.2\% & \cellcolor[HTML]{D9D9D9}37.54 / \textbf{20.30} & \cellcolor[HTML]{D9D9D9}15.8\% & \cellcolor[HTML]{D9D9D9}\textbf{77.45} / 60.42 & \cellcolor[HTML]{D9D9D9}3.7\% & \cellcolor[HTML]{D9D9D9}24.70 / 15.55 & \cellcolor[HTML]{D9D9D9}16.7\% & \cellcolor[HTML]{D9D9D9}75.61 / 76.92 & \cellcolor[HTML]{D9D9D9}1.8\% & \cellcolor[HTML]{D9D9D9}\textbf{8.0\%} \\
 & \multicolumn{2}{c}{MLP} & \cellcolor[HTML]{F2F2F2}79.10 / 56.45 & \cellcolor[HTML]{F2F2F2} & \cellcolor[HTML]{F2F2F2}32.73 / 14.81 & \cellcolor[HTML]{F2F2F2} & \cellcolor[HTML]{F2F2F2}72.65 / 57.89 & \cellcolor[HTML]{F2F2F2} & \cellcolor[HTML]{F2F2F2}24.42 / 16.09 & \cellcolor[HTML]{F2F2F2} & \cellcolor[HTML]{F2F2F2}69.18 / 75.43 & \cellcolor[HTML]{F2F2F2} & \cellcolor[HTML]{F2F2F2} \\
 & \multicolumn{2}{c}{CMoE,Hard} & 81.38 / 57.70 & 2.5\% & 30.01 / 12.89 & -10.6\% & 71.56 / 56.42 & -2.0\% & 22.05 / 15.69 & -6.1\% & 81.75 / 79.83 & 12.0\% & -0.8\% \\
\multirow{-3}{*}{16} & \multicolumn{2}{c}{CMoE,Soft} & \cellcolor[HTML]{D9D9D9}82.54 / 58.85 & \cellcolor[HTML]{D9D9D9}\textbf{4.3\%} & \cellcolor[HTML]{D9D9D9}\textbf{38.11} / 19.13 & \cellcolor[HTML]{D9D9D9}\textbf{22.8\%} & \cellcolor[HTML]{D9D9D9}71.99 / 59.99 & \cellcolor[HTML]{D9D9D9}1.4\% & \cellcolor[HTML]{D9D9D9}26.52 / 15.58 & \cellcolor[HTML]{D9D9D9}2.7\% & \cellcolor[HTML]{D9D9D9}76.51 / 75.99 & \cellcolor[HTML]{D9D9D9}5.7\% & \cellcolor[HTML]{D9D9D9}7.4\% \\
 & \multicolumn{2}{c}{MLP} & \cellcolor[HTML]{F2F2F2}79.23 / 56.50 & \cellcolor[HTML]{F2F2F2} & \cellcolor[HTML]{F2F2F2}33.50 / 16.13 & \cellcolor[HTML]{F2F2F2} & \cellcolor[HTML]{F2F2F2}72.04 / 58.78 & \cellcolor[HTML]{F2F2F2} & \cellcolor[HTML]{F2F2F2}18.67 / 15.42 & \cellcolor[HTML]{F2F2F2} & \cellcolor[HTML]{F2F2F2}71.67 / 72.69 & \cellcolor[HTML]{F2F2F2} & \cellcolor[HTML]{F2F2F2} \\
 & \multicolumn{2}{c}{CMoE,Hard} & 82.25 / 58.54 & 3.7\% & 30.56 / 11.97 & -17.3\% & 73.16 / 59.70 & 1.6\% & 23.22 / 15.58 & 12.7\% & \textbf{82.19} / 80.67 & \textbf{12.8\%} & 2.7\% \\
\multirow{-3}{*}{32} & \multicolumn{2}{c}{CMoE,Soft} & \cellcolor[HTML]{D9D9D9}82.39 / 57.18 & \cellcolor[HTML]{D9D9D9}2.6\% & \cellcolor[HTML]{D9D9D9}35.95 / 17.03 & \cellcolor[HTML]{D9D9D9}6.4\% & \cellcolor[HTML]{D9D9D9}70.14 / 59.66 & \cellcolor[HTML]{D9D9D9}-0.6\% & \cellcolor[HTML]{D9D9D9}\textbf{26.56} / 15.45 & \cellcolor[HTML]{D9D9D9}\textbf{21.2\%} & \cellcolor[HTML]{D9D9D9}77.61 / 77.18 & \cellcolor[HTML]{D9D9D9}7.2\% & \cellcolor[HTML]{D9D9D9}7.4\% \\
 & \multicolumn{2}{c}{MLP} & \cellcolor[HTML]{F2F2F2}79.35 / 57.23 & \cellcolor[HTML]{F2F2F2} & \cellcolor[HTML]{F2F2F2}35.22 / 17.95 & \cellcolor[HTML]{F2F2F2} & \cellcolor[HTML]{F2F2F2}72.46 / 56.65 & \cellcolor[HTML]{F2F2F2} & \cellcolor[HTML]{F2F2F2}22.29 / 14.90 & \cellcolor[HTML]{F2F2F2} & \cellcolor[HTML]{F2F2F2}71.12 / 75.84 & \cellcolor[HTML]{F2F2F2} & \cellcolor[HTML]{F2F2F2} \\
 & \multicolumn{2}{c}{CMoE,Hard} & 81.52 / 58.55 & 2.5\% & 31.93 / 12.28 & -20.5\% & 66.82 / 46.29 & -13.0\% & 23.88 / 15.85 & 6.7\% & 82.00 / 79.97 & 10.4\% & -2.8\% \\
\multirow{-3}{*}{64} & \multicolumn{2}{c}{CMoE,Soft} & \cellcolor[HTML]{D9D9D9}81.53 / 58.04 & \cellcolor[HTML]{D9D9D9}2.1\% & \cellcolor[HTML]{D9D9D9}35.64 / 18.81 & \cellcolor[HTML]{D9D9D9}3.0\% & \cellcolor[HTML]{D9D9D9}73.82 / 60.26 & \cellcolor[HTML]{D9D9D9}4.1\% & \cellcolor[HTML]{D9D9D9}25.89 / \textbf{16.77} & \cellcolor[HTML]{D9D9D9}14.3\% & \cellcolor[HTML]{D9D9D9}75.21 / 77.29 & \cellcolor[HTML]{D9D9D9}3.8\% & \cellcolor[HTML]{D9D9D9}5.5\% \\

\midrule
\multicolumn{14}{l}{\textbf{(e) Module generalization under LoRA-MoE setting (rank = 4)}} \\
\midrule

LLM  fine-tuning & \multicolumn{2}{c}{Connector\&Router} &  &  &  &  &  &  &  &  &  &  &  \\
LoRA & \multicolumn{2}{c}{MLP} & \cellcolor[HTML]{F2F2F2}80.70 / 56.42 & \cellcolor[HTML]{F2F2F2} & \cellcolor[HTML]{F2F2F2}36.35 / 16.32 & \cellcolor[HTML]{F2F2F2} & \cellcolor[HTML]{F2F2F2}64.34 / 57.02 & \cellcolor[HTML]{F2F2F2} & \cellcolor[HTML]{F2F2F2}\textbf{22.70} / \textbf{15.54} & \cellcolor[HTML]{F2F2F2} & \cellcolor[HTML]{F2F2F2}71.62 / 74.06 & \cellcolor[HTML]{F2F2F2} & \cellcolor[HTML]{F2F2F2} \\
LoRA-MoE & \multicolumn{2}{c}{MLP} & \textbf{85.17} / 61.29 & 7.1\% & 32.40 / 14.91 & -9.7\% & 78.68 / 65.77 & 18.8\% & 11.26 / 14.05 & -30.0\% & 76.66 / 78.67 & 6.6\% & -1.4\% \\
LoRA-MoE & \multicolumn{2}{c}{CMoE,Hard} & 84.10 / 61.25 & 6.4\% & 31.56 / 12.64 & -17.9\% & 78.58 / 62.51 & 15.9\% & 22.67 / 13.23 & -7.5\% & \textbf{80.65} / \textbf{79.89} & \textbf{10.2\%} & 1.4\% \\
LoRA-MoE & \multicolumn{2}{c}{CMoE,Soft} & \cellcolor[HTML]{D9D9D9}84.92 / \textbf{61.66} & \cellcolor[HTML]{D9D9D9}\textbf{7.3\%} & \cellcolor[HTML]{D9D9D9}\textbf{39.33} / \textbf{17.10} & \cellcolor[HTML]{D9D9D9}\textbf{6.5\%} & \cellcolor[HTML]{D9D9D9}\textbf{79.90} / \textbf{67.69} & \cellcolor[HTML]{D9D9D9}\textbf{21.4\%} & \cellcolor[HTML]{D9D9D9}18.73 / 13.75 & \cellcolor[HTML]{D9D9D9}-14.5\% & \cellcolor[HTML]{D9D9D9}78.90 / 77.98 & \cellcolor[HTML]{D9D9D9}7.7\% & \cellcolor[HTML]{D9D9D9}\textbf{5.7\%}
\\ \bottomrule
\end{tabular}
\end{table}

\paragraph{Evaluation metrics}
For ablation studies, we report BLEU-1 for the task of VQA, REG, and RG, IoU for the task of REC, Accuracy for the task of CLS.
In addition, we use $\Delta = \frac{1}{S}{\sum_{i=1}^{S}} \left(M_{m,i} - M_{b,i} \right) / M_{b,i} \times 100\%$ to evaluate the performance gains,
where $M_{m,i}$ and $M_{b,i}$ are the metrics of our model and baseline model,
$S$ can be the number of datasets or tasks.
For the overall comparison between models, we report more metrics such as F1, ROUGE, METEOR, RadGraph F1 and RadCliQ \citep{yu2023evaluating}.
See details at Appendix~\ref{D.1}.

\subsection{Ablation study}
\subsubsection{Ablation on module design}
\label{4.2.1}
\paragraph{Connector design}
Taking the connector of a two-layer MLP as baseline setup, we first discuss the performance of different multi-task learning hypothesis.
In Table~\ref{T2} (a), connectors based on conflict-synergy coexist hypothesis (CMoE with sparse / soft router) show a more holistic improvement trend in multi-task learning compared to connectors based on the conflict hypothesis (CMoE with hard router) and synergy hypothesis (linear, MLP, CMoE with constant router).
Though the hard router has a obvious lead on the CLS task, implying that the CLS task is better suited to a separate connector to avoid conflicts with other tasks.
The soft router achieves the best multi-task performance, indicating that it not only alleviates conflicts between tasks, but also promotes collaboration between tasks.
We then discuss three types of router strategy.
The strategy of combining token-level with task-level information is superior to using each information separately, indicating the effectiveness for considering the tug-of-war problem from both token and task level.

\paragraph{Resampler design}
We explore whether aggregating visual features through resampler has unfavorable effects in Table~\ref{T2} (b).
Despite an increase in compression rate $\alpha$ from 1 to 4, the performance of models utilizing projection aggregation is improved.
While the performance of average pooling and max pooling approaches is not satisfactory, especially the latter has severe performance degradation, which may be attributed to the excessive loss of feature information.
This phenomenon shows that appropriate visual feature compression can bring efficiency to the training process without losing or even improving performance.

\paragraph{Number of projection experts}
The number of projection experts $N$ is one of the most significant hyperparameters, which is closely related to the number of tasks and modalities that the CMoE module can accommodate.
It is a challenging study as the complexity of the scenario can end up overfitting to simpler tasks and modalities or underfitting complex ones.
As shown in Table~\ref{T2} (c), increasing the number of experts $N$,  namely an augmentation in parameters, still brings performance gains on some datasets, but the average gain tends to stabilize across all tasks and datasets.
Therefore, CMoE with 5 projection experts is sufficient to handle the tug-of-war problem in the existing medical multi-task learning scenarios and training configuration.
A higher value of $N$ does not bring the desired further improvement in total $\Delta$.

\subsubsection{Ablation on module generalization}
We demonstrate the generalization capability of the CMoE module in any configuration, especially when the key hyperparameters and strategies for LLM fine-tuning change.
We first focus on the rank of LoRA, which directly determines the LLM capacity, i.e., trainable parameters.
Our observations in Table~\ref{T2} (d) reveal that CMoE with soft router can steadily improve multi-task performance when LoRA rank increases from 4 to 64.
In Table~\ref{T2} (e), we introduce MoE to LoRA, namely LoRA-MoE, which is considered a favorable parameter-efficient tuning solution for multi-task applications \citep{liu2023moelora, chen2024llava}.
See details of LoRA-MoE at Appendix~\ref{A.2}.
We find that separate LoRA-MoE results in significant performance improvement in 3 tasks while degradation in 2 tasks, indicating that it does not achieve the efficient solution to the tug-of-war problem.
After combining CMoE with soft router, we achieve a balance of consistent performance gains, further demonstrating the necessity and effectiveness of mitigating the tug-of-war problem at the connector level in MLLMs.

\subsection{Interpretation}
\begin{figure}[!b]
  \centerline{\includegraphics[scale=0.425]{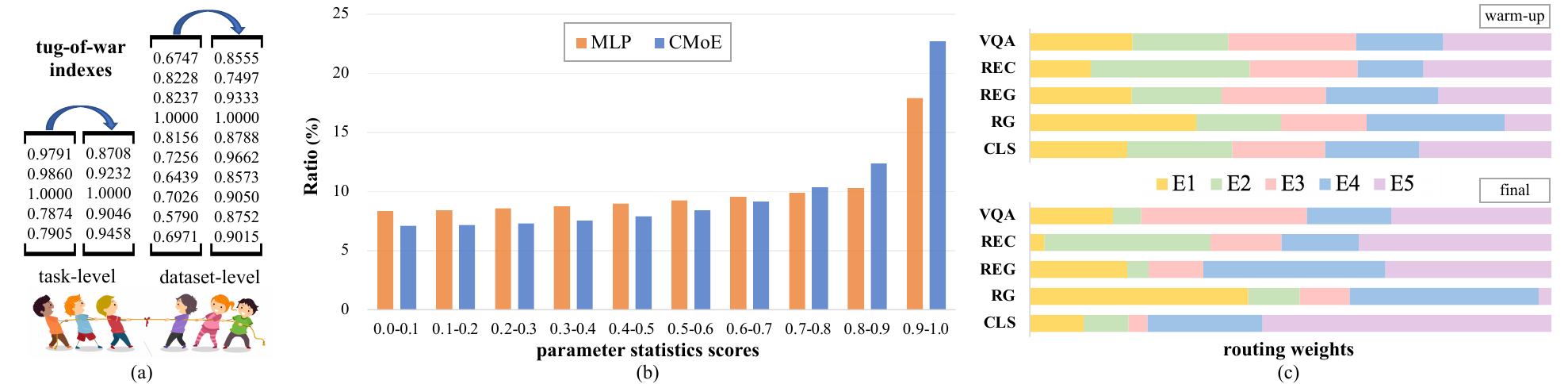}}
  \caption{Interpretation analysis of the tug-of-war problem. (a) changes in tug-of-war indexes, (b) changes in the distribution of parameter statistics scores,
  (c) routing weights for different tasks.}
\label{F4}
\end{figure}

We conduct interpretation analysis of the tug-of-war problem based on methods mentioned in Section~\ref{3.1.1}.
Specifically, we focus on the changes in the connector using CMoE compared to MLP and show how the tug-of-war problem is optimized:
(1) From the perspective of gradient optimization, we use maximum normalization to make the tug-of-war indexes comparable under different architectures.
CMoE results in a more consistent tug-of-war indexes, i.e. higher mean and smaller standard deviation, among different tasks or datasets, implying each individual gets a more balanced optimization, as shown in Figure~\ref{F4} (a).
(2) From the perspective of parameter statistics, we discrete the statistics scores into ten intervals and count the ratio of all parameters at connector by interval. 
CMoE results in an increase in the proportion of high-value intervals in Figure~\ref{F4} (b).
We show the routing weights of projection experts after the warm-up stage and the final model in Figure~\ref{F4} (c).
CMoE adaptively learns different patterns of routing weights for different tasks.

\begin{figure}[!b]
  \centerline{\includegraphics[scale=0.47]{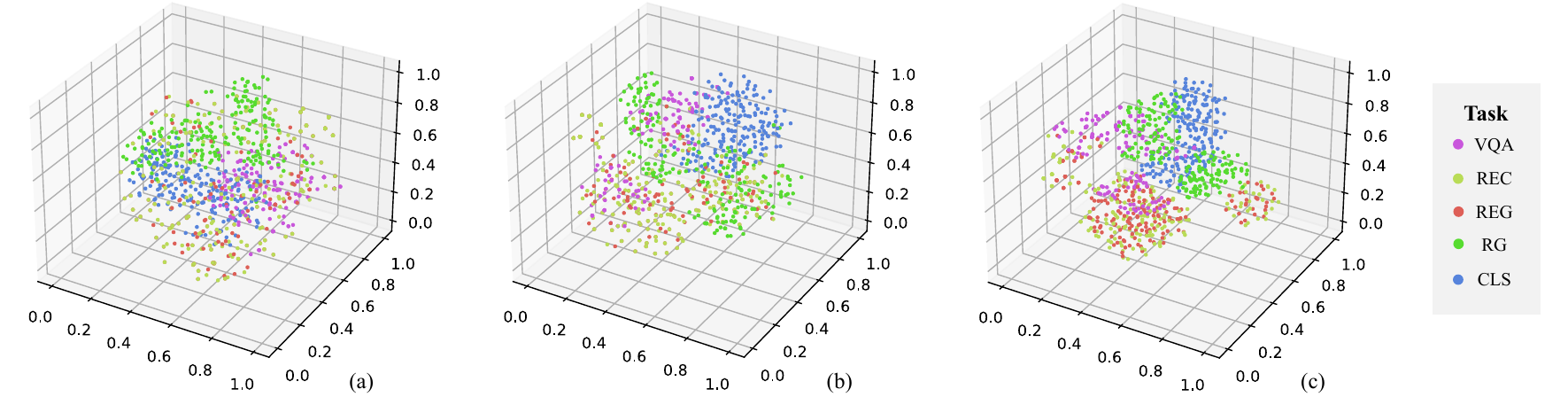}}
  \caption{Visual features distribution maps-3D. (a) $\bm{f}_{v}^{ag}$ distribution, (b) $\bm{f}_{v}^{align}$ distribution obtained through MLP, (c) $\bm{f}_{v}^{align}$ distribution obtained through CMoE.}
\label{F5}
\end{figure}

To better reflect the coexistence of conflict and synergy among tasks, as well as the critical role played by the connector, we visualize the distribution of visual features before and after passing through the connector using the t-SNE method \citep{van2008visualizing}.
From the perspective of multi-task learning, we randomly select 200 samples from each task.
It can be observed that CMoE promotes the optimization of the tug of war problem when aligning the visual space with the textual space of the LLM in Figure~\ref{F5}.
Specifically, visual features of the same task are more tightly distributed.
For fine-grained REC and REG tasks, the distribution is highly overlapping, which facilitates synergy between tasks.
For coarse-grained CLS task, the distribution is significantly different from other tasks, which is consistent with the conclusion in Section~\ref{4.2.1}.
We also provide visualization analysis of visual features on different medical image modalities in Appendix~\ref{D.4}

\begin{table}[!b]
\scriptsize
\tabcolsep= 0.24cm
\renewcommand\arraystretch{1.1}
\caption{Model capability comparison with open source medical MLLMs.
The mean and standard deviation of performance of Uni-Med are obtained after several 300k iterations.
Results with \textbf{bold}, \uline{underlines} and gray background are the overall best, second, and zero-shot performance, respectively.}
\label{T3}
\centering
\begin{tabular}{c c c c c c c c}
\toprule
\textbf{Task} & \textbf{Dataset} & \textbf{Metric} & \textbf{Med-Flamingo} & \textbf{RadFM} & \textbf{LLaVA-Med} & \textbf{XrayGPT} & \textbf{Uni-Med} \\
\midrule
 &  & BLEU-1 & \cellcolor[HTML]{F2F2F2}21.51 & {\ul 81.66} & 76.95 & - & \textbf{82.12}$\pm$0.38 \\
 & \multirow{-2}{*}{Slake-VQA} & F1 & \cellcolor[HTML]{F2F2F2}23.66 & {\ul 82.38} & 77.30 & - & \textbf{83.07}$\pm$0.34 \\ \cmidrule(r){2-8}
 &  & BLEU-1 & \cellcolor[HTML]{F2F2F2}33.38 & \cellcolor[HTML]{F2F2F2}24.83 & {\ul 46.42} & - & \textbf{58.07}$\pm$0.32 \\
\multirow{-4}{*}{\begin{tabular}[c]{@{}c@{}}Visual\\ Question\\ Answering\end{tabular}} & \multirow{-2}{*}{Path-VQA} & F1 & \cellcolor[HTML]{F2F2F2}34.01 & \cellcolor[HTML]{F2F2F2}25.20 & {\ul 47.08} & - & \textbf{58.74}$\pm$0.33 \\ \midrule
 &  & BLEU-1 & \cellcolor[HTML]{F2F2F2}23.25 & 6.81 & \cellcolor[HTML]{F2F2F2}19.90 & {\ul  27.11} & \textbf{27.79}$\pm$2.50 \\
 &  & BLEU-4 & \cellcolor[HTML]{F2F2F2}1.92 & 1.52 & \cellcolor[HTML]{F2F2F2}0.59 & {\ul 3.56} &  \; \textbf{6.46}$\pm$0.20 \\
 &  & ROUGE-1 & \cellcolor[HTML]{F2F2F2}18.73 & 16.81 & \cellcolor[HTML]{F2F2F2}15.65 & {\ul 24.35} & \textbf{28.81}$\pm$1.22 \\
 &  & ROUGE-2 & \cellcolor[HTML]{F2F2F2}2.28 & 4.48 & \cellcolor[HTML]{F2F2F2}1.13 & {\ul 4.97} &\;  \textbf{9.62}$\pm$0.99 \\
 &  & ROUGE-L & \cellcolor[HTML]{F2F2F2}12.25 & 12.67 & \cellcolor[HTML]{F2F2F2}10.29 & {\ul 16.29} & \textbf{22.58}$\pm$2.86\\
 &  & METEOR & \cellcolor[HTML]{F2F2F2}7.95 & 5.32 & \cellcolor[HTML]{F2F2F2}5.47 & {\ul 9.71} & \textbf{10.59}$\pm$0.87 \\
 &  & RadGraph-F1 & \cellcolor[HTML]{F2F2F2}7.15 & 7.19 & \cellcolor[HTML]{F2F2F2}2.86 & {\ul 9.00} & \textbf{13.98}$\pm$2.45 \\
 &  & RadCliQ-v0$\downarrow$ & \cellcolor[HTML]{F2F2F2}4.44 & 4.43 & \cellcolor[HTML]{F2F2F2}4.79 & {\ul 4.42} & \textbf{3.75}$\pm$0.17 \\
 & \multirow{-9}{*}{MIMIC-CXR} & RadCliQ-v1$\downarrow$ & \cellcolor[HTML]{F2F2F2}1.80 & 1.82 & \cellcolor[HTML]{F2F2F2}2.03 & {\ul 1.79} & \textbf{1.38}$\pm$0.11 \\ \cmidrule(r){2-8}
 &  & BLEU-1 & \cellcolor[HTML]{F2F2F2}8.14 & - & \cellcolor[HTML]{F2F2F2}{\ul 9.46} & \cellcolor[HTML]{F2F2F2}8.51 & {\textbf{15.80}$\pm$0.24} \\
 &  & BLEU-4 & \cellcolor[HTML]{F2F2F2}0.45 & - & \cellcolor[HTML]{F2F2F2}{\ul 0.59} & \cellcolor[HTML]{F2F2F2}0.23 & {\; \textbf{2.47}$\pm$0.08} \\
 &  & ROUGE-1 & \cellcolor[HTML]{F2F2F2}{\ul 11.37} & - & \cellcolor[HTML]{F2F2F2}11.31 & \cellcolor[HTML]{F2F2F2}8.00 & {\textbf{14.32}$\pm$0.03} \\
 &  & ROUGE-2 & \cellcolor[HTML]{F2F2F2}0.93 & - & \cellcolor[HTML]{F2F2F2}{\ul 1.02} & \cellcolor[HTML]{F2F2F2}0.45 & {\; \textbf{2.68}$\pm$0.01} \\
 &  & ROUGE-L & \cellcolor[HTML]{F2F2F2}{\ul 9.65} & - & \cellcolor[HTML]{F2F2F2}8.96 & \cellcolor[HTML]{F2F2F2}6.48 & {\textbf{12.29}$\pm$0.04} \\
 &  & METEOR & \cellcolor[HTML]{F2F2F2}4.31 & - & \cellcolor[HTML]{F2F2F2}{\ul 5.51} & \cellcolor[HTML]{F2F2F2}3.60 & {\; \textbf{5.92}}$\pm$0.07 \\
 &  & RadGraph-F1 & \cellcolor[HTML]{F2F2F2}1.85 & - & \cellcolor[HTML]{F2F2F2}{\ul 2.63} & \cellcolor[HTML]{F2F2F2}1.32 & {\; \textbf{4.91}$\pm$0.31} \\
 &  & RadCliQ-v0$\downarrow$ & \cellcolor[HTML]{F2F2F2}4.00 & - & \cellcolor[HTML]{F2F2F2}{\ul 3.88} & \cellcolor[HTML]{F2F2F2}4.17 & {\; \textbf{3.59}$\pm$0.02} \\
\multirow{-18}{*}{\begin{tabular}[c]{@{}c@{}}Report\\ Generation\end{tabular}} & \multirow{-9}{*}{MPx-Single} & RadCliQ-v1$\downarrow$ & \cellcolor[HTML]{F2F2F2}1.62 & - & \cellcolor[HTML]{F2F2F2}{\ul 1.55} & \cellcolor[HTML]{F2F2F2}1.72 & {\; \textbf{1.37}}$\pm$0.01 \\ \midrule
 & DermaMNIST & Accuracy & \cellcolor[HTML]{F2F2F2}1.15 & \cellcolor[HTML]{F2F2F2}{\ul 5.14} & - & - & \textbf{76.96}$\pm$0.46 \\ \cmidrule(r){2-8}
\multirow{-2}{*}{\begin{tabular}[c]{@{}c@{}}Image\\ Classification\end{tabular}} & OrganMNIST & Accuracy & \cellcolor[HTML]{F2F2F2}8.90 & \cellcolor[HTML]{F2F2F2}{\ul 18.90} & - & - & \textbf{78.07}$\pm$1.63 \\
\bottomrule
\end{tabular}
\end{table}

\subsection{Overall comparison}
To demonstrate the model capabilities of Uni-Med on multi-task learning,
four open source and state-of-the-art medical MLLMs including Med-Flamingo \citep{moor2023med}, RadFM \citep{wu2023towards}, LLaVA-Med \citep{li2024llava}, and XrayGPT \citep{thawkar2023xraygpt} are used for performance comparison in Table~\ref{T3}.
Any method of fine-tuning will inevitably lead to changes in the initial capability of the model.
Therefore, we use readily available model checkpoints for testing, following the prompt template requirements of different models.
Under this comparison strategy, if the training datasets of a model and Uni-Med intersect and strictly follow the official partition, it is fair and comparable to Uni-Med on these datasets.
Specifically, LLaVA-MED provides dataset-specific fine-tuning checkpoints on Slake-VQA and Path-VQA separately.
XrayGPT focuses on the task of report generation and utilizes MIMIC-CXR as training dataset.
RadFM provides a model checkpoint for joint fine-tuning on Slake-VQA, MIMIC-CXR and MPx-Single.
However, we do not list performance of RadFM on MPx-Single as we have identified the issue of data leakage, see Appendix~\ref{D.2}.

The results in Table~\ref{T3} show that our Uni-Med achieves leading and competitive evaluation metrics across all tasks, which has the following prominent advantages:
(1) Uni-Med is able to handle a greater variety of medical tasks, which is attributed to multi-task learning during training process.
Due to the fact that the above MLLMs do not support input and output in coordinate form, we report the performance of Uni-Med on REC and REG tasks at Appendix~\ref{D.5}.
Based on the different input and output forms supported by each model, we have also listed the zero-shot results in Table~\ref{T3} for reference only.
(2) Uni-Med achieves better results through joint training fine-tuning rather than dataset-specific fine-tuning like LLaVA-Med, which benefits from efficient optimization of the tug-of-war problem.
In addition to directly compare the capability of existing models, we take LLaVA-Med as an example to compare the capability of model architectures in Appendix~\ref{D.6}.

\section{Conclusion}
In this paper, we present a novel open-source medical generalist foundation model Uni-Med, which can handle six different medical tasks including question answering, visual question answering, report generation, referring expression comprehension, referring expression generation and image classification.
Benefiting from the proposed CMoE, which combines MoE with the connector, Uni-Med achieves efficient solution to the tug-of-war problem in multi-task learning.
Uni-Med not only achieves competitive or superior performance compared to the open-source state-of-the-art medical MLLMs, but also provides interpretability analysis from the perspective of gradient optimization and parameter statistics on how the tug-of-war problem is optimized.
We hope Uni-Med can greatly promote the development of medical generalist foundation models and inspire more research toward generalist medical artificial intelligence.

\section{Limitations}
While Uni-Med has demonstrated strong potential as a unified and generalist medical foundation model, it still exhibits several limitations:
(1) Limitations in handling genuine 3D medical image inputs.
Most commonly used medical image are in 3D.
Same as most medical MLLMs, we process 3D images into 2D slices as input, resulting in significant information loss.
(2) The potential of performance gains in more complex multi-modal and multi-task learning scenarios has not yet been explored. 
Uni-Med use 12 datasets of 6 medical tasks, with a total data volume of 140k.
(3) The potential of performance gains in different LLM backbones has not yet been explored. 
Uni-Med utilizes LLaMA2-7B.
(4) Deeper theoretical analysis of tug of war problem remains to be explored.
We attempt to combine the existing methods to analyze it from the perspective gradient optimization and parameter statistics.
(5) Potential negative societal impacts.
We cannot prevent potential malicious or unintended uses, such as generating fake profiles or wrong medical diagnoses, and provide necessary safeguards.

{
\small
\nocite{*}
\bibliographystyle{named}
\bibliography{neurips24}
}

\newpage
\appendix

\section{Component design}
\subsection{Type of the routing network}
\label{A.1}
\paragraph{Constant router}
The simplest routing network is to assign equal weights to the output of each expert, which can be expressed as:
\begin{small}
\begin{equation}
R_{\textit{constant}}(x_i)=\{1/N\}_{k=1}^N
\end{equation}
\end{small}

\paragraph{Hard router}
Each token is assigned to a specific expert based on its type (task / modal), with the number of experts being equal to the number of token types. It can be formulated as:
\begin{small}
\begin{equation}
\begin{gathered}
R_{\textit {hard }}\left(x_i\right)=\left\{\text { IsType }\left(x_i, k\right)\right\}_{k=1}^N \\
\text { IsType }\left(x_i, k\right)= \begin{cases}1, & \text { if } x_i \text { belongs to type } k \\
0, & \text { otherwise }\end{cases}
\end{gathered}
\end{equation}
\end{small}

\paragraph{Sparse router}
Using a small network $g$, the sparse router computes a score vector for each token, with a length equal to the number of experts $N$. Subsequently, the Top-$K$ function retains the top-$K$ values in the vector, while setting all other values to zero. Finally, the \textit{Softmax} function is applied to obtain the final routing vector. The whole process is shown as follows:
\begin{small}
\begin{equation}
\begin{gathered}
R_{\textit {sparse }}\left(x_i\right)=\textit{Softmax}\left(\text{Top-$K$}\left(g\left(x_i\right), K\right)\right) \\
\text{Top-$K$}(v, K)= \begin{cases}v, & \text { if $v$ is in the top } K \\
0, & \text { otherwise }\end{cases}
\end{gathered}
\end{equation}
\end{small}

\paragraph{Soft Router}
Similar to the sparse router, the soft router computes a score vector for each token through a small network $g$. Subsequently, it applies the \textit{Sigmoid} function to the score vector and normalizes it, yielding the final routing vector. It can be formulated as:
\begin{small}
\begin{equation}
R_{\textit{soft}}(x_i)=\frac{\textit{Sigmoid}(g(x_i))}{\textit{Sum}(\textit{Sigmoid}(g(x_i)))}
\end{equation}
\end{small}

\subsection{LoRA-MoE}
\label{A.2}
LoRA-MoE freezes the original parameters of the model to preserve world knowledge and introduces LoRA experts to learn new knowledge, thereby improving performance across multiple downstream tasks with few parameters.

Specifically, given a frozen linear layer with a weight matrix $W_0 \in \mathbf{R}^{d_{\textit{in}} \times d_{\textit{out}}}$, LoRA-MoE creates $N$ low-rank trainable matrix pairs $A_k$ and $B_k$, where $A_k \in \mathbf{R}^{d_{\textit{in}} \times r}$, $B_k \in \mathbf{R}^{r \times d_{\textit{out}}}$, and the rank $r \ll min(d_{\textit{in}}, d_{out})$. As in the case of LoRA, $A_k$ is initialized with a random Gaussian distribution, and $B_k$ is initialized to zero. During training, the parameters of $W_0$ are frozen, and the parameters of $A_k$ and $B_k$ are updated. The forward process of a LoRA-MoE layer can be represented as:
\begin{small}
\begin{equation}
h=W_0 x_i+ \Delta W x_i=W_0 x_i+\frac{\alpha}{r}\sum_{k=1}^N R(x_i) A_k B_k x_i
\end{equation}
\end{small}
where $x_i$ is the input token, $R$ is the router in the LoRA-MoE layer, $\alpha$ is the learning rate scaling factor, and $h$ is the output token.
In ablation experiments, we transform each linear layer in the LLM into a LoRA-MoE layer with a sparse router. The rank $r=4$, the learning rate scaling factor $\alpha=8$, the number of LoRA experts $N=5$, and select the top 2 experts.

\section{Dataset}
\label{B}
\subsection{Data source}

\paragraph{MedQA} MedQA \citep{jin2021disease} is a open-domain multiple-choice question answering dataset for solving medical problems. These questions are sourced from professional medical board exams, which feature diverse content and typically demand a comprehensive understanding of related medical concepts learned from medical textbooks in order to provide accurate answers. This dataset covers three languages: English, simplified Chinese, among which there are 12,723 QA pairs for English.

\paragraph{PubMedQA} PubMedQA \citep{jin2019pubmedqa} is a biomedical question answering dataset collected from PubMed abstracts. The task of PubMedQA is to answer research questions with yes/no/maybe using the corresponding abstracts. It has 1K expert-annotated, 61.2K unlabeled and 211.3K artificially generated QA instances. Each instance consists of: (1) a question which is either an existing research article title or derived from one, (2) a context which is the corresponding abstract without its conclusion,(3) a long answer, which is the conclusion of the abstract and, presumably, answers the research question, and (4) a yes/no/maybe answer which summarizes the conclusion.

\paragraph{Slake-VQA} Slake-VQA \citep{liu2021slake} is a semantically annotated, knowledge-enhanced bilingual (English and Chinese) VQA dataset for radiology images. It contains 642 annotated images accompanied by 14,028 question-answer pairs, spanning 12 diseases, 39 organ systems, and 3 imaging modalities (CT, MRI, and X-ray). Questions are either open-ended (free-form) or closed-ended (balanced yes/no) related to various aspects of the image content such as plane, quality, position, organ, abnormality, size, color, shape, and knowledge graph.

\paragraph{Path-VQA} Path-VQA \citep{he2020pathvqa} is a pathology VQA dataset comprising 4,998 pathology images and 32,799 question-answer pairs. These pathology images are sourced from medical textbooks and online digital libraries. Each image is associated with multiple QA pairs pertaining to different aspects of the pathology including color, location, appearance, shape, etc. The dataset includes 16,465 open-ended questions, which make up 50.2\% of the total and are categorized into six types: what, where, when, whose, how, and how much/how many. The remaining questions are close-ended "yes/no" questions, with a balanced distribution of 8,145 "yes" answers and 8,189 "no" answers. In the official dataset split, the training set, validation set and test set contain 19,755, 6,279 and 6,761 QA pairs, respectively.

\paragraph{SA-Med2D-20M} SA-Med2D-20M \citep{ye2023sa} is a large-scale segmentation dataset of 2D medical images built upon numerous public and private datasets. It consists of 4.6 million 2D medical images and 19.7 million corresponding masks, covering almost the whole body and showing significant diversity. It comprises 10 modalities, with CT and MR modalities being predominant in both the number of images and masks. Specifically, there are 2338,753 images and 12547,037 masks for CT and 2217,633 images and 7147,784 masks for MR. This is primarily attributed to their widespread presence in public medical image segmentation datasets and the 3D dimension of CT and MR scans, which yields a high volume of slices when segmented across three axes.

\paragraph{MIMIC-CXR} MIMIC-CXR \citep{johnson2019mimic} is a large dataset of chest radiographs with free-text radiology reports. A total of 377,110 images are available in the dataset from 227,835 image studies collected for 65,379 patients. Each patient may have multiple studies and each study may contain one or more images associated with the same free-text report. Images in MIMIC-CXR are collected from multiple view positions: e.g., anterior-posterior (AP), posterior- anterior, and lateral (LA). Protected health information (PHI) in radiology reports and images is removed, which results in missing information in some sentences of the reports.

The MIMIC-CXR-JPG dataset is derived from MIMIC-CXR, providing JPG format files derived from the DICOM images and structured labels derived from the free-text reports. The aim of MIMIC-CXR-JPG is to provide a convenient processed version of MIMIC-CXR, as well as to provide a standard reference for data splits and image labels.

RadFM \citep{wu2023towards} processes radiology reports in MIMIC-CXR by extracting the indication, findings, and impression sections, and removing redundant white spaces. Images without reports and reports where the findings section can not be extracted are discarded from both the training and test sets. Additionally, reports with findings sections exceeding 800 characters are filtered out. To enhance the model's capability to process images from different view positions, images of different orientations associated with the same report are treated as independent samples.

\paragraph{MPx} MPx \citep{wu2023towards} is a report generation dataset collected from the MedPix website (https://medpix.nlm.nih.gov/) and organized by cases. Each case includes multiple radiologic scans, general clinical findings, discussions, and diagnostic results. Additionally, MPx provides scan-level annotations, such as image modality, shooting plane, and captions for each scan. The dataset is divided into MPx-Single and MPx-Multi, with annotations provided at the case level and scan level, respectively.

\paragraph{MedMNIST v2} MedMNIST v2 \citep{yang2023medmnist} is a large-scale MNIST-like collection of standardized biomedical images, including 2D datasets with resolutions up to 224×224 pixels and 3D datasets with resolutions up to 64×64×64 voxels. The 2D datasets include 12 subsets: PathMNIST, ChestMNIST, DermaMNIST, OCTMNIST, PneumoniaMNIST, RetinaMNIST, BreastMNIST, BloodMNIST, TissueMNIST, OrganAMNIST, OrganCMNIST, and OrganSMNIST. The 3D datasets comprise 6 subsets: OrganMNIST3D, NoduleMNIST3D, FractureMNIST3D, AdrenalMNIST, VesselMNIST3D, and SynapseMNIST3D. Covering primary data modalities in biomedical images, it is designed to perform classification on lightweight 2D and 3D images with various data scales (from 100 to 100,000) and diverse tasks (binary/multi-class, ordinal regression and multi-label). The comprehensive dataset, comprising approximately 708K 2D images and 10K 3D images, supports a wide range of research and educational purposes in biomedical image analysis, computer vision, and machine learning.

DermaMNIST, a 2D subset of MedMNIST v2, is based on HAM10000 \citep{tschandl2018ham10000, codella2019skin}, a large collection of multi-source dermatoscopic images of common pigmented skin lesions. Comprising 10,015 dermatoscopic images, the dataset is categorized into 7 distinct classes: actinic keratoses and intraepithelial carcinoma, basal cell carcinoma, benign keratosis-like lesions, dermatofibroma, melanoma, melanocytic nevi, and vascular lesions.

OrganSMNIST, another 2D subset of MedMNIST v2, is based on 3D computed tomography (CT) images from Liver Tumor Segmentation Benchmark (LiTS) \citep{bilic2023liver}. Organ labels are obtained by using bounding-box annotations of 11 body organs from another study \citep{xu2019efficient}. Hounsfield-Unit (HU) of the 3D images are transformed into grey scale with a abdominal window. Subsequently, 2D images are cropped from the center slices of the 3D bounding boxes in sagittal views. Comprising 25,211 images, the dataset is categorized into 11 distinct classes: bladder, left femur, right femur, heart, left kidney, right kidney, liver, left lung, right lung, pancreas, and spleen.

\paragraph{Custom dataset splitting} 
To prevent the model from encountering training images during testing, the official dataset split from Slake-VQA is not utilized. Instead, we randomly divide all images into training and testing sets at a ratio of 6:1, along with their respective QA pairs and bounding boxes. Consequently, the training set comprises 550 images, 6018 English QA pairs, and 1421 bounding boxes, while the testing set includes 92 images, 1014 English QA pairs, and 201 bounding boxes.

For MIMIC-CXR, JPG images provided in MIMIC-CXR-JPG and the corresponding reports from RadFM are used for the report generation task. The training set is a subset of the original training set, containing 9,997 samples, while the test set remains the same as the original test set, containing 3,858 samples.

\subsection{Well-crafted datasets for REC and REG tasks}
\begin{figure}
  \centerline{\includegraphics[scale=0.48]{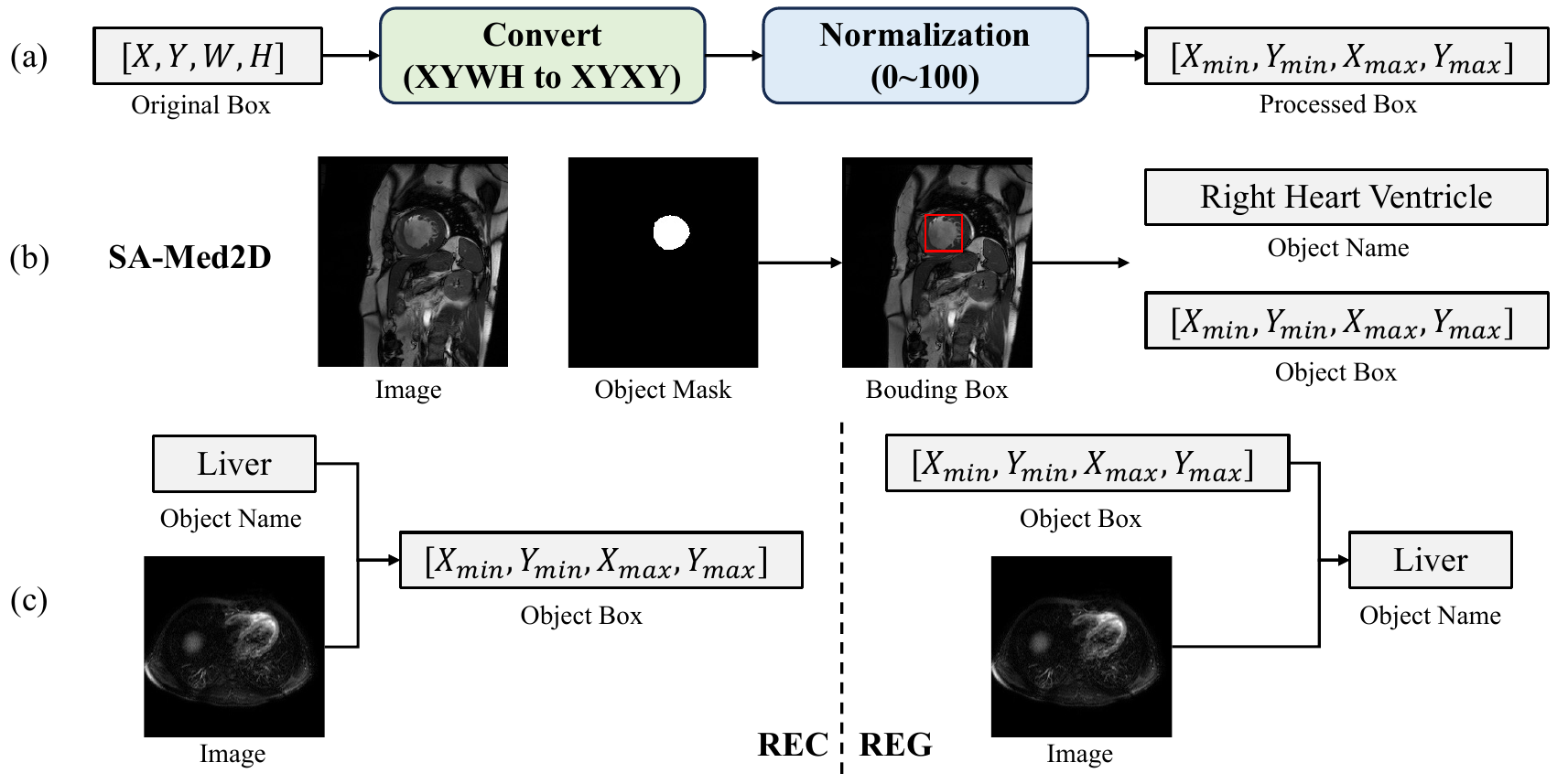}}
  \caption{Data production process for REC and REG tasks. (a) the process of transforming bounding boxes in Slake-VQA, (b) the process of obtaining bounding boxes from masks in SA-Med2D, (c) the input-output organization of REC and REG tasks.}
\label{F6}
\end{figure}

\paragraph{Slake-REC / Slake-REG} As a semantically-labeled knowledge-enhanced dataset for medical visual question answering, Slake-VQA provides bounding boxes for each object in the image. As shown in Figure~\ref{F6} (a), the original format of each bounding box is $[X, Y, W, H]$. First, we convert it to the $[X_{min}, Y_{min}, X_{max}, Y_{max}]$ format. Assuming the relative size of each image is 100×100, we then normalize each coordinate value in the bounding box to fall within the range of 0 to 100.

As shown in Figure~\ref{F6} (c), in the REC task, an image and object name are given to find the object's bounding box. In the REG task, an image and object bounding box are provided to identify the object's name. The Slake-REC and Slake-REG datasets are thus created.

\paragraph{SA-Med2D-REC / SA-Med2D-REG} Each image in the SA-Med2D-20M dataset has one or more masks, with each mask corresponding to an object. As shown in Figure~\ref{F6} (b), we calculate the bounding box for each mask and normalize it to a range of 0 to 100, resulting in a bounding box for each object in the $[X_{min}, Y_{min}, X_{max}, Y_{max}]$ format.

The SA-Med2D-REC and SA-Med2D-REG datasets are organized as depicted in Figure~\ref{F6} (c). 10,000 samples each are selected from the CT and MR subsets as the training set, and 2,000 samples each are selected as the test set.

\subsection{Data availability}

In the Table \ref{T4}, we list the links for each dataset, the number of samples in the training and test sets, and their licenses.

\begin{table}[h]
\scriptsize
\label{T4}
\centering
\renewcommand\arraystretch{1.4}
\caption{Data availability.}
\begin{tabular}{cccc}
\toprule
\textbf{Dataset} & \multicolumn{1}{c}{\textbf{Link}} & \textbf{Train / Test Split} & \textbf{License} \\ \midrule
MedQA & https://github.com/jind11/MedQA & 10178 / 1273 & MIT License \\
PubMedQA & https://github.com/pubmedqa/pubmedqa & 500 / 500 & MIT License \\
Slake-VQA &  & 6018 / 1014 & Open Access \\
Slake-REC &  & 1421 / 201 & - \\
Slake-REG & \multirow{-3}{*}{https://www.med-vqa.com/slake} & 1421 / 201 & - \\
Path-VQA & https://github.com/UCSD-AI4H/PathVQA & 19755 / 6761 & MIT License \\
SA-Med2D-20M & & - & Apache-2.0 license \\
SA-Med2D-REC & & 20000 / 4000 & - \\
SA-Med2D-REG & \multirow{-3}{*}{https://openxlab.org.cn/datasets/GMAI/SA-Med2D-20M} & 20000 / 4000 & - \\
 & {https://physionet.org/content/mimic-cxr-jpg/2.1.0} &  & \begin{tabular}[c]{@{}c@{}}PhysioNet Credentialed\\ Health Data License 1.5.0\end{tabular} \\
\multirow{-2}{*}{MIMIC-CXR} & {https://huggingface.co/datasets/chaoyi-wu/RadFM\_data\_csv} & \multirow{-2}{*}{9997 / 3858} & Apache-2.0 license \\
 & {https://huggingface.co/datasets/chaoyi-wu/MedPix-Images} &  & Open Access \\
\multirow{-2}{*}{MPx-Single} & https://huggingface.co/datasets/chaoyi-wu/RadFM\_data\_csv & \multirow{-2}{*}{31416 / 6664} & Apache-2.0 license \\
DermaMNIST & https://medmnist.com & 7007 / 2005 & Apache-2.0 License \\
OrganSMNIST & https://medmnist.com & 13932 / 8827 & Apache-2.0 License \\
\bottomrule
\end{tabular}
\end{table}

\section{Multi-task instruction template}
\label{C}

We have designed different instruction templates for different datasets. During the training process, when a sample from a dataset is selected, an instruction template is also sampled from the corresponding dataset's template pool and used to format the sample. Examples of instruction templates for each dataset are shown below.

\begin{tcolorbox}[title = {MedQA},breakable]
\begin{small}
    \textbf{Example 1}: [qa] A researcher evaluates healthy breast tissue from 100 women, 50 women that were pregnant at the time of the study and 50 age-matched non-pregnant women. The breast tissue in pregnant women contained an increased number of acinar glands with epithelial proliferation compared to the non-pregnant women. Which process caused this change?
\end{small}
    \tcblower
\begin{small}
    \textbf{Example 2}: [qa] If you are a doctor, please answer the following question briefly: a researcher evaluates healthy breast tissue from 100 women, 50 women that were pregnant at the time of the study and 50 age-matched non-pregnant women. The breast tissue in pregnant women contained an increased number of acinar glands with epithelial proliferation compared to the non-pregnant women. Which process caused this change?
\end{small}
\end{tcolorbox}

\begin{tcolorbox}[title = {PubMedQA},breakable]
\begin{small}
    \textbf{Example 1}: [qa] Does the severity of obstructive sleep apnea predict patients requiring high continuous positive airway pressure?
\end{small}
    \tcblower
\begin{small}
    \textbf{Example 2}: [qa] If you are a doctor, please answer the following question using "yes", "no" or "maybe": does the severity of obstructive sleep apnea predict patients requiring high continuous positive airway pressure?
\end{small}
    
\end{tcolorbox}

\begin{tcolorbox}[title = {Slake-VQA / Path-VQA},breakable]
\begin{small}
\textbf{Example 1}: <Img> <ImageFeature> </Img> [vqa] What modality is used to take this image?
\end{small}
\tcbline
\begin{small}
\textbf{Example 2}: <Img> <ImageFeature> </Img> [vqa] Based on the image, respond to this question with a short answer: what modality is used to take this image?
\end{small}
\end{tcolorbox}

\begin{tcolorbox}[title = {Slake-REC / SA-Med2D-REC},breakable]
\begin{small}
\textbf{Example 1}: <Img> <ImageFeature> </Img> [refer] Liver.
\end{small}
\tcbline
\begin{small}
\textbf{Example 2}: <Img> <ImageFeature> </Img> [refer] Give me the location of liver.
\end{small}
\tcbline
\begin{small}
\textbf{Example 3}: <Img> <ImageFeature> </Img> [refer] Where is liver?
\end{small}
\tcbline
\begin{small}
\textbf{Example 4}: <Img> <ImageFeature> </Img> [refer] From this image, tell me the location of liver.
\end{small}
\tcbline
\begin{small}
\textbf{Example 5}: <Img> <ImageFeature> </Img> [refer] The location of liver is
\end{small}
\tcbline
\begin{small}
\textbf{Example 6}: <Img> <ImageFeature> </Img> [refer] Could you tell me the location for liver?
\end{small}
\tcbline
\begin{small}
\textbf{Example 7}: <Img> <ImageFeature> </Img> [refer] Where can I locate the liver?
\end{small}
\end{tcolorbox}

\begin{tcolorbox}[title = {Slake-REG / SA-Med2D-REG},breakable]
\begin{small}
\textbf{Example 1}: <Img> <ImageFeature> </Img> [identify] {<16><36><42><61>}
\end{small}
\tcbline
\begin{small}
\textbf{Example 2}: <Img> <ImageFeature> </Img> [identify] What object is in this location {<16><36><42><61>}?
\end{small}
\tcbline
\begin{small}
\textbf{Example 3}: <Img> <ImageFeature> </Img> [identify] Identify the object present at this location {<16><36><42><61>}.
\end{small}
\tcbline
\begin{small}
\textbf{Example 4}: <Img> <ImageFeature> </Img> [identify] What is it in {<16><36><42><61>}?
\end{small}
\tcbline
\begin{small}
\textbf{Example 5}: <Img> <ImageFeature> </Img> [identify] Describe this object in {<16><36><42><61>}.
\end{small}
\tcbline
\begin{small}
\textbf{Example 6}: <Img> <ImageFeature> </Img> [identify] This {<16><36><42><61>} is
\end{small}
\tcbline
\begin{small}
\textbf{Example 7}: <Img> <ImageFeature> </Img> [identify] The object in {<16><36><42><61>} is
\end{small}
\end{tcolorbox}

\begin{tcolorbox}[title = {MIMIC-CXR},breakable]
\begin{small}
\textbf{Example 1}: <Img> <ImageFeature> </Img> [caption] Describe the given chest x-ray image in detail.
\end{small}
\tcbline
\begin{small}
\textbf{Example 2}: <Img> <ImageFeature> </Img> [caption] Take a look at this chest x-ray and describe the findings and impression.
\end{small}
\tcbline
\begin{small}
\textbf{Example 3}: <Img> <ImageFeature> </Img> [caption] Could you provide a detailed description of the given x-ray image?
\end{small}
\tcbline
\begin{small}
\textbf{Example 4}: <Img> <ImageFeature> </Img> [caption] Describe the given chest x-ray image as detailed as possible.
\end{small}
\tcbline
\begin{small}
\textbf{Example 5}: <Img> <ImageFeature> </Img> [caption] What are the key findings in this chest x-ray image?
\end{small}
\end{tcolorbox}

\begin{tcolorbox}[title = {MPx-Single},breakable]
\begin{small}
\textbf{Example 1}: <Img> <ImageFeature> </Img> [caption] Describe this input image.
\end{small}
\tcbline
\begin{small}
\textbf{Example 2}: <Img> <ImageFeature> </Img> [caption] Help captioning the image.
\end{small}
\tcbline
\begin{small}
\textbf{Example 3}: <Img> <ImageFeature> </Img> [caption] What can be inflected from the scan?
\end{small}
\tcbline
\begin{small}
\textbf{Example 4}: <Img> <ImageFeature> </Img> [caption] Can you give a caption for this image?
\end{small}
\tcbline
\begin{small}
\textbf{Example 5}: <Img> <ImageFeature> </Img> [caption] Can you provide a brief summary of the radiology image?
\end{small}
\tcbline
\begin{small}
\textbf{Example 6}: <Img> <ImageFeature> </Img> [caption] Please write a report about the image?
\end{small}
\tcbline
\begin{small}
\textbf{Example 7}: <Img> <ImageFeature> </Img> [caption] Can you provide an analysis of this image?
\end{small}
\tcbline
\begin{small}
\textbf{Example 8}: <Img> <ImageFeature> </Img> [caption] Can you explain what is shown in this image?
\end{small}
\tcbline
\begin{small}
\textbf{Example 9}: <Img> <ImageFeature> </Img> [caption] What can be indicated from the radiologic scans?
\end{small}
\tcbline
\begin{small}
\textbf{Example 10}: <Img> <ImageFeature> </Img> [caption] What can you infer from this photograph?
\end{small}
\end{tcolorbox}

\begin{tcolorbox}[title = {DermaMNIST},breakable]
\begin{small}
\textbf{Example}: <Img> <ImageFeature> </Img> [cls] Which category does this multi-source dermatoscopic image of common pigmented skin lesions belong to: actinic keratoses and intraepithelial carcinoma, basal cell carcinoma, benign keratosis-like lesions, dermatofibroma, melanoma, melanocytic nevi, or vascular lesions?
\end{small}
\end{tcolorbox}

\begin{tcolorbox}[title = {OrganSMNIST},breakable]
\begin{small}
\textbf{Example}: <Img> <ImageFeature> </Img> [cls] Which category does this CT image belong to: bladder, left femur, right femur, heart, left kidney, right kidney, liver, left lung, right lung, pancreas, or spleen?
\end{small}
\end{tcolorbox}

\section{Experiments}
\subsection{Evaluation metrics}
\label{D.1}
\paragraph{F1 Score}
Assuming $m$ is the number of common words in the candidate $C$ and the reference $R$ with the number of words of $c$ and $r$, the precision and recall for a candidate sentence can be calculated as:
\begin{small}
\begin{equation}
\textit{precision}=\frac{m}{c}
\end{equation}
\end{small}
\begin{small}
\begin{equation}
\textit{recall}=\frac{m}{r}
\end{equation}
\end{small}

Considering class imbalance, F1 score is used to evaluate the performance of the model on both the VQA and REG tasks, which means the harmonic mean of precision and recall.
A higher average F1 score for the dataset indicates a higher performance of the model.
\begin{small}
\begin{equation}
\text{F1}=\frac{2 \times \textit{precision} \times \textit{recall}}{\textit{precision}+\textit{recall}}
\end{equation}
\end{small}

\paragraph{BLEU-N}
We use BLEU-1 to assess the model's performance on both the VQA and REG tasks, while employing both BLEU-1 and BLEU-4 to evaluate its performance in the report generation task. Given the candidate $C$ and reference $R$, BLEU-N is defined as:
\begin{small}
\begin{equation}
\text{BLEU-N}=\frac{\sum_{\mathrm{gram}_N \in C}Count_{clip}(\mathrm{gram}_N)}{\sum_{\mathrm{gram}_N \in C}Count(\mathrm{gram}_N)}
\end{equation}
\end{small}

When N=1, the above formula calculates BLEU-1; when N=4, it calculates BLEU-4.

\paragraph{ROUGE-N}
We use ROUGE-1 and ROUGE-2 to evaluate the performance of the model on the RG task.
Given the candidate $C$ and reference $R$, ROUGE-N is defined as:
\begin{small}
\begin{equation}
\text{ROUGE-N}=\frac{\sum_{\mathrm{gram}_N \in R}Count_{match}(\mathrm{gram}_N)}{\sum_{\mathrm{gram}_N \in R}Count(\mathrm{gram}_N)}
\end{equation}
\end{small}

When N=1, the above formula calculates ROUGE-1; when N=2, it calculates ROUGE-2.

\paragraph{ROUGE-L}
ROUGE-L is also used to evlaute the quality of the generated text on the task of report generation, which stands for recall-oriented understudy for gisting evaluation with the longest common subsequence. Given the candidate $C$ and reference $R$, let $LCS(C, R)$ be the length of the longest common subsequence, which is determined by using dynamic programming, it can be an defined as:
\begin{small}
\begin{equation}
\text{ROUGE-L}=\frac{(1+\beta^2)R_{LCS}P_{LCS}}{R_{LCS}+\beta^2P_{LCS}}
\end{equation}
\end{small}

where $R_{LCS}=\frac{LCS(C,R)}{L_C}$, $P_{LCS}=\frac{LCS(C,R)}{L_R}$, $\beta=\frac{P_{LCS}}{R_{LCS}}$. $L_C$ and $L_R$ represent the length of the candidate and reference. 
A higher ROUGE-L score means that the generated text shares more of the same sequences of words as the reference text, which typically indicates better quality in terms of capturing the salient points of the reference.

\paragraph{METEOR}
METEOR is also used to evlaute the quality of the generated text on the task of report generation, which stands for metric for evaluation of translation with explicit ordering. METEOR for a sentence is computed as:
\begin{small}
\begin{equation}
\text{METEOR}=(1-p) \times \frac{\textit{precision} \times \textit{recall}}{\alpha \times \textit{precision}+(1-\alpha) \times \textit{recall}}
\end{equation}
\end{small}

where $p=\gamma(\frac{ch}{m})^\theta$ is the penalty factor. $ch$ is the number of chunks, which means a contiguous ordered block. $\alpha$, $\theta$ and $\gamma$ are hyperparameters determined according to different datasets.

\paragraph{RadGraph F1}
To assess the semantic accuracy in the task of report generation, RadGraph F1 computes the overlap in clinical entities and relations between a machine-generated report and a radiologist-generated report.
Specifically, following the criteria in RadGraph \citep{jain2021radgraph}, two entities are matched if their tokens (words in the original report) and labels (entity type) match.
Two relations are matched if their start and end entities match and the relation type matches.
RadGraph F1 metric computes the overlap in entities and relations separately and reports their average.

\paragraph{RadCliQ}
RadCliQ (radiology report clinical quality) is also used to assess the semantic accuracy in the task of report generation.
Two versions of the RadCliQ metric: RadCliQ-v0 and RadCliQ-v1 both use a machine learning model to take in values from other metrics, such as BERTScore and CheXbert vector similarity, and then produce a composite score based on these input values, which predict the total number of errors in a report.

\paragraph{IoU}
We use IoU (Intersection over Union) to evaluate the performance of the model on the REC task.
It can be formulated as:
\begin{small}
\begin{equation}
\text{IoU}=\frac{P \cap G}{P \cup G}
\end{equation}
\end{small}

where $P$ is the prediction area of the model, $G$ is the area of the ground truth.

\paragraph{R@0.5}
We alse use R@0.5 to evaluate the performance of the model on the referring expression comprehension task. R stands for recall, and 0.5 denotes the IoU threshold. When the IoU between the prediction and the ground truth is greater than or equal to 0.5, it is considered a true positive (TP). When the IoU is less than 0.5, it is considered a false negative (FN). Therefore, for a sample with only one bounding box, R@0.5 can be formalized as:
\begin{small}
\begin{equation}
\text{R@0.5}=\frac{TP}{TP+FN}=\begin{cases}1, & \text{IoU} \geq 0.5 \\
0, & \text{IoU} < 0.5 \end{cases}
\end{equation}
\end{small}

\subsection{Data leakage issue of RadFM on MPx-Single}
\label{D.2}
When we directly use the model checkpoint provided by RadFM open-source repository for model inference,
we find that the model outputs for many samples were completely consistent with ground truth.
This issue only occurs on the MPx-Single dataset, while we strictly follow the official test set split.
This appears to be unreasonable, raising suspicions of potential data leakage. Here are some examples:

\begin{tcolorbox}[title = {Example 1}]
    \begin{minipage}{0.35\textwidth}
        \includegraphics[width=0.9\textwidth]{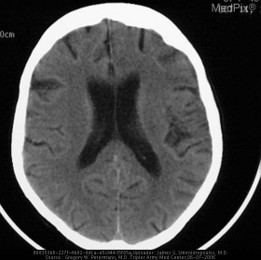}
    \end{minipage}%
    \begin{minipage}{0.65\textwidth}
    \textbf{Instruction}: Describe this input image.
    \newline
    \newline
    \textcolor{red}{\textbf{Ground truth}: Neuroradiology Quiz}
    \newline
    \newline
    \textcolor{red}{\textbf{RadFM}: Neuroradiology Quiz}
    \newline
    \newline
    \textbf{Uni-Med}: axial noncontrast ct image shows a large, well-circumscribed, non-aggressive appearing mass in the right frontal lobe.
    \end{minipage}
\end{tcolorbox}

\begin{tcolorbox}[title = {Example 2}]
    \begin{minipage}{0.35\textwidth}
        \includegraphics[width=0.9\textwidth]{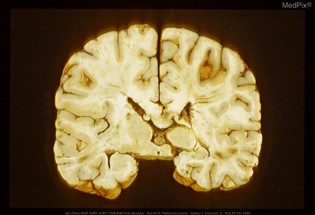}
    \end{minipage}%
    \begin{minipage}{0.65\textwidth}
    \textbf{Instruction}: Describe this input image.
    \newline
    \newline
    \textcolor{red}{\textbf{Ground truth}: Focal necrosis of splenium of corpus callosum secondary to subacute diffuse axonal injury.}
    \newline
    \newline
    \textcolor{red}{\textbf{RadFM}: Focal necrosis of splenium of corpus callosum secondary to subacute diffuse axonal injury.}
    \newline
    \newline
    \textbf{Uni-Med}: there is a large, well-circumscribed, extra-axial mass in the right frontal region. the mass is predominantly cystic with a few foci of solid tissue. there is no evidence of hemorrhage or enhancement on the mr sequences.
    \end{minipage}
\end{tcolorbox}

\begin{tcolorbox}[title = {Example 3}]
    \begin{minipage}{0.35\textwidth}
        \includegraphics[width=0.9\textwidth]{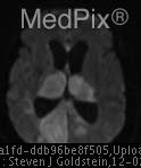}
    \end{minipage}%
    \begin{minipage}{0.65\textwidth}
    \textbf{Instruction}: Describe this input image.
    \newline
    \newline
    \textcolor{red}{\textbf{Ground truth}: MRI day 2- hydrocephalus Acute bilateral cerebellar infarcts Acute bilateral thalamic infarcts Acute right occipital lobe infarct MRA no flow in distal basilar artery or posterior cerebal arteries.}
    \newline
    \newline
    \textcolor{red}{\textbf{RadFM}: MRI day 2- hydrocephalus Acute bilateral cerebellar infarcts Acute bilateral thalamic infarcts Acute right occipital lobe infarct MRA no flow in distal basilar artery or posterior cerebal arteries.}
    \newline
    \newline
    \textbf{Uni-Med}: acute right mca infarct. acute infarction of the right cerebellar hemisphere. acute infarction of the right brainstem. acute cerebral edema.
    \end{minipage}
\end{tcolorbox}

\subsection{Ablation on special token and identifier}
\label{D.3}
We have designed vision-level special task tokens and text-level special task identifiers for visual features and text prompt, respectively.
Through ablation experiment, we verify whether they have a positive effect on model performance.
As shown in Table~\ref{T5}, we observe that text-level special task identifiers bring limited improvement.
In contrast, vision-level special task tokens significantly improve the model's overall performance on all datasets, further illustrating the effectiveness of mitigating the tug-of-war problem at the connector.

\begin{table}[h]
\tiny
\tabcolsep= 0.068cm
\renewcommand\arraystretch{1.1}
\caption{Ablation Experiments on special token and identifier.}
\label{T5}
\centering
\begin{tabular}{ccccccccccccccccccc}
\toprule
 & \multicolumn{2}{c}{Special Token / Identifier} & \multicolumn{2}{c}{VQA} &  & \multicolumn{2}{c}{REC} &  & \multicolumn{2}{c}{REG} &  & \multicolumn{2}{c}{RG} &  & \multicolumn{2}{c}{CLS} &  &  \\
\multirow{-2}{*}{Connector} & Text-level & Vision-level & \multicolumn{2}{c}{BLEU-1} & \multirow{-2}{*}{Avg.} & \multicolumn{2}{c}{IoU} & \multirow{-2}{*}{$\Delta \left(\uparrow\right)$} & \multicolumn{2}{c}{BLEU-1} & \multirow{-2}{*}{$\Delta \left(\uparrow\right)$} & \multicolumn{2}{c}{BLEU-1} & \multirow{-2}{*}{$\Delta \left(\uparrow\right)$} & \multicolumn{2}{c}{Accuracy} & \multirow{-2}{*}{$\Delta \left(\uparrow\right)$} & \multirow{-2}{*}{\begin{tabular}[c]{@{}c@{}}Total\\ $\Delta \left(\uparrow\right)$\end{tabular}} \\ \midrule
MLP & - & - & \cellcolor[HTML]{F2F2F2}79.81 & \cellcolor[HTML]{F2F2F2}56.48 & \cellcolor[HTML]{F2F2F2} & \cellcolor[HTML]{F2F2F2}35.18 & \cellcolor[HTML]{F2F2F2}16.26 & \cellcolor[HTML]{F2F2F2} & \cellcolor[HTML]{F2F2F2}74.54 & \cellcolor[HTML]{F2F2F2}58.42 & \cellcolor[HTML]{F2F2F2} & \cellcolor[HTML]{F2F2F2}18.55 & \cellcolor[HTML]{F2F2F2}15.50 & \cellcolor[HTML]{F2F2F2} & \cellcolor[HTML]{F2F2F2}76.26 & \cellcolor[HTML]{F2F2F2}73.64 & \cellcolor[HTML]{F2F2F2} & \cellcolor[HTML]{F2F2F2} \\ \midrule
 & - & - & 81.59 & 57.35 & 1.9\% & 36.76 & 18.74 & 9.9\% & 76.07 & 58.81 & 1.4\% & 24.71 & 15.42 & 16.4\% & 74.46 & 76.07 & 0.5\% & 6.0\% \\
 & - & \CheckmarkBold & 81.33 & 57.29 & 1.7\% & \textbf{37.85} & 20.14 & 15.7\% & 77.23 & \textbf{62.72} & \textbf{5.5\%} & 23.29 & \textbf{15.74} & 13.6\% & \textbf{76.76} & 76.55 & 2.3\% & 7.8\% \\
 & \CheckmarkBold & - & \textbf{81.79} & 57.69 & \textbf{2.3\%} & 35.51 & 17.79 & 5.2\% & 74.43 & 61.34 & 2.4\% & \textbf{26.27} & 15.61 & \textbf{21.2\%} & 76.56 & \textbf{77.21} & \textbf{2.6\%} & 6.7\% \\
\multirow{-4}{*}{CMoE} & \CheckmarkBold & \CheckmarkBold & 81.52 & \textbf{57.75} & 2.2\% & 37.54 & \textbf{20.30} & \textbf{15.8\%} & \textbf{77.45} & 60.42 & 3.7\% & 24.70 & 15.55 & 16.7\% & 75.61 & 76.92 & 1.8\% & \textbf{8.0\%}\\
\bottomrule
\end{tabular}
\end{table}

\subsection{Visualization analysis of visual features on image modalities}
\label{D.4}
\begin{figure}
  \centerline{\includegraphics[scale=0.395]{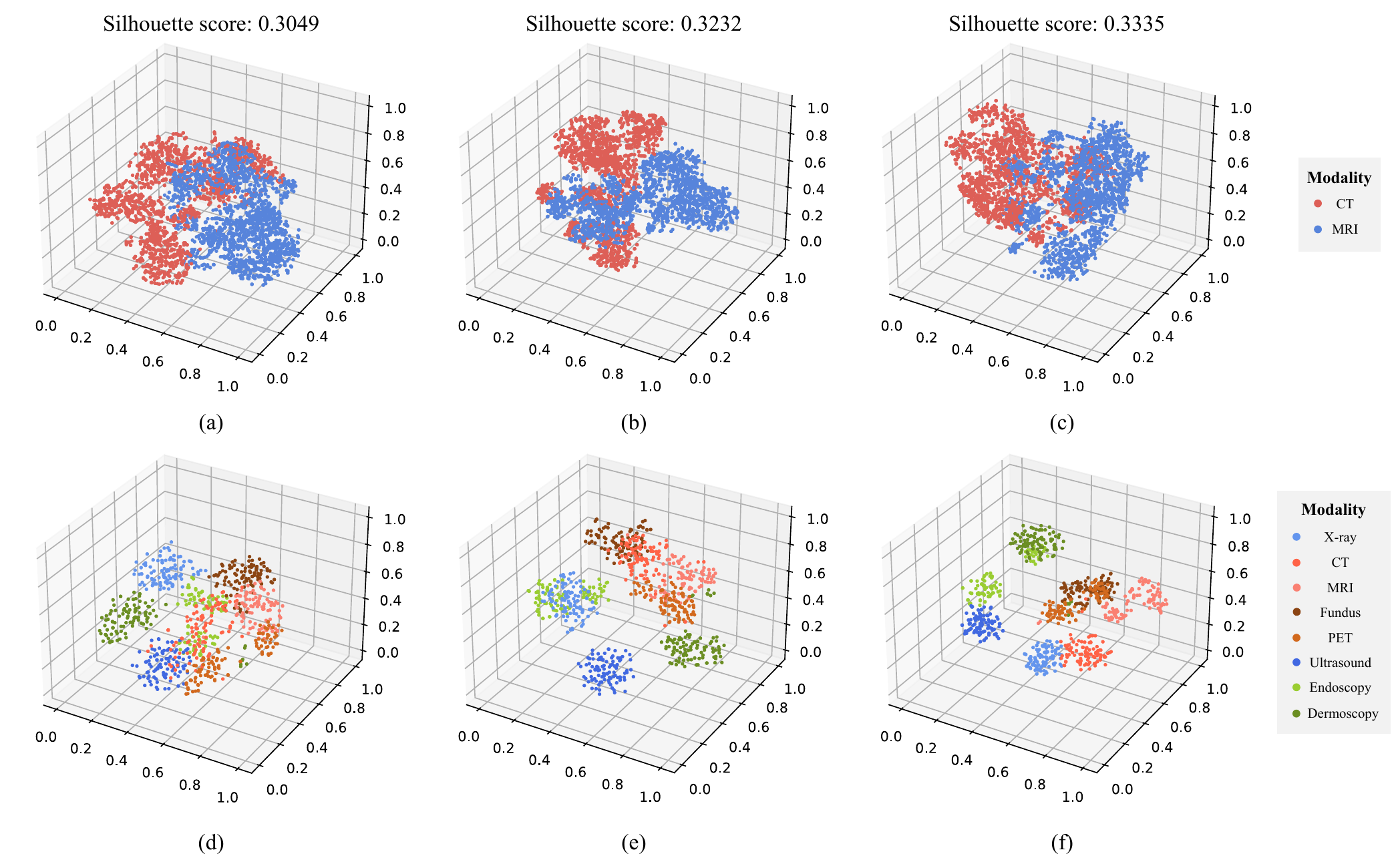}}
  \caption{Visual features distribution on image modalities. (a)-(c) The feature distribution of CT and MRI modalities in the REG task. (a) passing through the frozen visual encoder.(b) passing through the MLP connector. (c) passing through the CMoE. (d)-(f) The feature distribution of 8 modalities after passing through the frozen visual encoder. (d) EVA-CLIP ViT-G/14. (e) CLIP ViT-L/14. (f) BiomedCLIP ViT-B/16.}
\label{F7}
\end{figure}

We use t-SNE to visualize the distribution of visual features by modalities in Figure~\ref{F7}. 
We first observe the visual feature distribution of different modalities under the same task in Figure~\ref{F7} (a-c). 
The feature of CT and MRI modalities in the REG task already have good discriminability after passing through the frozen visual encoder.
After passing through the connector, the improvement in Silhouette score (from 0.3049 to 0.3335) is relatively limited.
In addition, we select 100 samples from each of the 8 modalities and observe their visual feature distributions after passing through different visual encoders in Figure~\ref{F7} (d-f). 
It can still be observed that visual features of different modalities already have specific patterns in the feature space, whether using EVA-CLIP, CLIP or BiomedCLIP.

The above findings also provide an explanation for why we attempt to explicitly introduce task information instead of modality information in CMoE.
When aligning visual and language embedding spaces through the connector in Uni-Med's multi-modal and multi-task scenario, task information is more difficult to distinguish than modality information.

\subsection{Performance of Uni-Med on REC and REG tasks}
\label{D.5}
We report the metrics of Uni-Med on the tasks of referring expression comprehension and referring exression generation in Table~\ref{T6}.
The mean and standard deviation of performance of Uni-Med are obtained after several 300k iterations.

\begin{table}[h]
\scriptsize
\tabcolsep= 0.4cm
\renewcommand\arraystretch{1.1}
\caption{Performance of Uni-Med on REC and REG tasks.}
\label{T6}
\centering
\begin{tabular}{c c c c}
\toprule
\textbf{Task} & \textbf{Dataset} & \textbf{Metric} & \multicolumn{1}{c}{\textbf{Uni-Med}} \\ \midrule
\multirow{4}{*}{Referring Expression Comprehension} & \multirow{2}{*}{Slake-REC} & IoU & 37.71$\pm$0.52 \\
 &  & R@0.5 & 39.30$\pm$0.76 \\ \cmidrule(r){2-4}
 & \multirow{2}{*}{SA-Med2D-REC} & IoU & 21.60$\pm$2.19 \\
 &  & R@0.5 & 14.42$\pm$3.20 \\ \midrule
\multirow{6}{*}{Referring Expression Generation} & \multirow{3}{*}{Slake-REG} & BLEU-1 & 75.78$\pm$2.05 \\ 
 &  & F1 & 77.35$\pm$1.97 \\
 &  & Accuracy & 68.16$\pm$1.32 \\ \cmidrule(r){2-4}
 & \multirow{3}{*}{SA-Med2D-REG} & BLEU-1 & 61.47$\pm$1.76 \\
 &  & F1 & 62.17$\pm$1.90 \\
 &  & Accuracy & 57.69$\pm$1.07 \\ \bottomrule
\end{tabular}
\end{table}

\subsection{Comparison of architecture capability between Uni-Med and LLaVA-Med}
\label{D.6}
In addition to directly compare the capability of existing models, we take LLaVA-Med as an example to compare the capability of model architectures.

Specifically, we use the checkpoints of the second stage (medical instruction tuning) to perform two strategies of LLM full parameter fine-tuning:
(1) Dataset-specific fine-tuning;
(2) Joint training fine-tuning.
The data split and the prompt format are completely consistent with Uni-Med and LLaVA-Med, respectively. 
Both strategies last for 3 epochs (the same as Uni-Med).
The results are shown in Table~\ref{T7}.

Following the model architecture of LLaVA-Med, there is a serious tug-of-war problem when we implement joint fine-tuning strategy on multiple tasks and datasets.
While the strategy of dataset-specific fine-tuning has significantly improved the evaluation metrics of each dataset.

It is worth noting that Uni-Med has achieved competitive and leading results through joint training, without dataset-specific fine-tuning.
It can be concluded that the model architecture of Uni-Med, especially the design of CMoE, has achieved a superior solution to the tug-of-war problem, which reduces interference and promotes more efficient knowledge sharing.

\begin{table}[!ht]
\scriptsize
\caption{Comparison of architecture capability between Uni-Med and LLaVA-Med. We utilize dataset-specific fine-tuning and joint training fine-tuning on LLaVA-Med, respectively.}
\renewcommand\arraystretch{1.1}
    \centering
    \begin{tabular}{c c c c c c}
\toprule
\multirow{2}{*}{\textbf{Task}} & \multirow{2}{*}{\textbf{Dataset}} & \multirow{2}{*}{\textbf{Metric}} & \multicolumn{2}{c}{\textbf{LLaVA-Med}}  & \textbf{Uni-Med} \\ 
        & & & \textbf{Joint Training} & \textbf{Dataset-specific} & \textbf{Joint Training} \\ \midrule
        \multirow{4}{*}{Visual Question Answering} & \multirow{2}{*}{Slake-VQA} & BLEU-1 & 33.69  & 72.00  & \textbf{82.12}  \\ 
        ~ & ~ & F1 & 35.83  & 73.07  & \textbf{83.07}  \\ \cmidrule(r){2-6}
        ~ & \multirow{2}{*}{Path-VQA} & BLEU-1 & 37.79 & 56.86  & \textbf{58.07}  \\ 
        ~ & ~ & F1 & 38.55 & 57.51  & \textbf{58.74}  \\ \midrule
        \multirow{12}{*}{Report Generation} & \multirow{6}{*}{MIMIC-CXR} & BLEU-1 & 20.43 & 21.03  & \textbf{27.79}  \\ 
        ~ & ~ & BLEU-4 & 4.86 & 4.96  & \textbf{6.46}  \\ 
        ~ & ~ & ROUGE-1 & 26.11 & 28.28  & \textbf{28.81}  \\ 
        ~ & ~ & ROUGE-2 & 7.66 & 9.01  & \textbf{9.62}  \\ 
        ~ & ~ & ROUGE-L & 19 & 20.61  & \textbf{22.58}  \\ 
        ~ & ~ & METEOR & 8.73 & 8.89  & \textbf{10.59}  \\ \cmidrule(r){2-6}
        ~ & \multirow{6}{*}{MPx-Single} & BLEU-1 & 15.11 & 14.63  & \textbf{15.80}  \\ 
        ~ & ~ & BLEU-4 & 2.4 & 1.75  & \textbf{2.47}  \\ 
        ~ & ~ & ROUGE-1 & 13.22 & 13.03  & \textbf{14.32}  \\ 
        ~ & ~ & ROUGE-2 & 2.39 & 2.19  & \textbf{2.68}  \\ 
        ~ & ~ & ROUGE-L & 10.99 & 10.85  & \textbf{12.29}  \\ 
        ~ & ~ & METEOR & 5.83 & 5.79  & \textbf{5.92}  \\ \midrule
        \multirow{2}{*}{Image Classification} & DermaMNIST & Accuracy & 25.84  & \textbf{79.95}  & 76.96  \\ \cmidrule(r){2-6}
        ~ & OrganSMNIST & Accuracy & 66.80  & 77.84  & \textbf{78.07}  \\ \midrule
        \multirow{4}{*}{Referring Expression Comprehension} & \multirow{2}{*}{Slake-REC} & IoU & 4.07  & 22.41  & \textbf{37.71}  \\
        ~ & ~ & R@0.5 & 1.99  & 18.41  & \textbf{39.30}  \\ \cmidrule(r){2-6}
        ~ & \multirow{2}{*}{SA-Med2D-REC} & IoU & 8.64  & 17.67  & \textbf{21.60}  \\ 
        ~ & ~ & R@0.5 & 4.75  & 9.98  & \textbf{14.42}  \\  \midrule
        \multirow{6}{*}{Referring Expression Generation} & \multirow{3}{*}{Slake-REG} & BLEU-1 & 27.21  & 50.79  & \textbf{75.78}  \\ 
        ~ & ~ & F1 & 30.97  & 53.15  & \textbf{77.35}  \\ 
        ~ & ~ & Accuracy & 20.40  & 44.78  & \textbf{68.16}  \\ \cmidrule(r){2-6}
        ~ & \multirow{3}{*}{SA-Med2D-REG} & BLEU-1 & 45.83  & 55.15  & \textbf{61.47}  \\ 
        ~ & ~ & F1 & 47.11  & 55.98  & \textbf{62.17}  \\ 
        ~ & ~ & Accuracy & 40.80  & 50.92  & \textbf{57.69} \\ \bottomrule
    \end{tabular}
\label{T7}
\end{table}

\end{document}